\documentclass[11pt]{article}

\usepackage[final]{acl}

\usepackage{times}
\usepackage{latexsym}

\usepackage[T1]{fontenc}

\usepackage[utf8]{inputenc}

\usepackage{microtype}

\usepackage{inconsolata}

\usepackage{graphicx}
\usepackage{array}
\usepackage{hyperref}
\usepackage{url}
\usepackage{amsmath}  
\usepackage{amsthm}  
\usepackage{amssymb}
\newtheorem{problem}{Problem}
\usepackage{graphicx} 
\usepackage[capitalise]{cleveref}
\usepackage{booktabs}
\usepackage{multirow} 
\usepackage{caption}
\usepackage{subcaption}

\usepackage{xspace}
\newcommand{\methodFont}{\texttt}

\newcommand{\ours}{\methodFont{MATU}\xspace}

\title{Every Response Counts: Quantifying Uncertainty of LLM-based Multi-Agent Systems through Tensor Decomposition}

\author{
  Tiejin Chen$^1$\quad 
  Huaiyuan Yao$^1$\quad 
  Jia Chen$^2$\quad 
  Evangelos E. Papalexakis$^2$\quad 
  Hua Wei$^1$ \\
  $^1$Arizona State University \\
  $^2$University of California, Riverside
}

\begin{document}
\maketitle
\footnotetext[1]{\texttt{tiejin@asu.edu}}
\begin{abstract}
While Large Language Model-based Multi-Agent Systems (MAS) consistently outperform single-agent systems on complex tasks, their intricate interactions introduce critical reliability challenges arising from communication dynamics and role dependencies. Existing Uncertainty Quantification methods, typically designed for single-turn outputs, fail to address the unique complexities of the MAS. Specifically, these methods struggle with three distinct challenges: the cascading uncertainty in multi-step reasoning, the variability of inter-agent communication paths, and the diversity of communication topologies. To bridge this gap, we introduce \ours, a novel framework that quantifies uncertainty through tensor decomposition. \ours moves beyond analyzing final text outputs by representing entire reasoning trajectories as embedding matrices and organizing multiple execution runs into a higher-order tensor. By applying tensor decomposition, we disentangle and quantify distinct sources of uncertainty, offering a comprehensive reliability measure that is generalizable across different agent structures. We provide comprehensive experiments to show that \ours effectively estimates holistic and robust uncertainty across diverse tasks and communication topologies.

\end{abstract}

\section{Introduction}
While multi-agent systems (MAS), where multiple LLM-based agents collaborate, consistently outperform single-agent systems on complex tasks, their complex interactions introduce critical and MAS-specific reliability challenges~\citep{li2023camel,wu2024autogen,wang2025anymac,zhang2024g}. Uncertainty in these systems emerges not from a single agent's isolated error, but from the complex dynamics of communication, role dependencies, and consensus-building. A minor, early mistake can irrevocably cascade through the collaboration. Therefore, Uncertainty Quantification~(UQ) for MAS is critical, especially when MAS is applied to domains such as scientific discovery~\citep{lu2024ai}, education~\cite{yao-etal-2026-instructional}, healthcare decision support~\citep{kim2024mdagents,tang2023medagents}, and transportation~\citep{da2024open,li2024urbangpt,yao2025comal}. For example, in a medical context, an initial agent's misdiagnosis can steer the entire pipeline toward a confidently asserted but dangerously flawed treatment plan and a reliable UQ method could help to mitigate such risks.

Uncertainty estimation itself is not a new concern in machine learning. For decades, it has been a fundamental part of supervised learning tasks such as regression~\citep{ye2024uncertainty,amini2020deep} and classification~\citep{gal2016dropout,sensoy2018evidential}. However, the landscape changes dramatically in the era of Large Language Models~(LLMs) and their deployment as agents. Unlike traditional supervised tasks, LLMs must generate free-form text. This generative nature introduces new uncertainty factors that go beyond classical classification or regression. Recent work has proposed specialized UQ methods for LLMs~\cite{liu2025uncertainty,xia2025survey}, focusing on semantic consistency such as semantic entropy~\citep{kuhn2023semantic} and graph-based methods~\citep{lin2023generating, da2024llm,da2025understanding}. All these methods rely on natural language inference (NLI) models~\citep{maccartney2009natural} to capture the similarity between the answers. While these techniques have proven useful, most of them concentrate on single-turn outputs from a standalone model.

In contrast, the setting of LLM-based agents, especially multi-agent systems, raises a new class of challenges: (1)~\textit{Multi-step reasoning}: Many current UQ frameworks measure uncertainty by assessing outputs' semantic consistency with NLI models~\citep{kuhn2023semantic,lin2023generating}. This approach fails in the context of multi-step reasoning for two key reasons. First, applying it only to the final output ignores the rich uncertainty information embedded in the reasoning process. Second, a naive attempt to fix this by concatenating entire reasoning trajectories into long documents makes it difficult for NLI models, which are typically designed for sentence-pair tasks and have context limitations. More importantly, in MAS, uncertainty is distributed across heterogeneous agents; an NLI model cannot distinguish whether a contradiction arises from an individual agent’s hallucination or a logical misalignment between two different agents during a handoff.  (2)~\textit{Inter-agent communication diversity}: For the same query, agents may collaborate through different sequences of interactions across runs. However, UQ methods that focus on semantic diversity are blind to this path diversity. (3)~\textit{Communication topology diversity}: Existing UQ methods are designed and validated for single models, which represent a fixed computational structure. In the MAS ecosystem, however, systems are built with diverse communication topologies. The effectiveness of a UQ method developed for a single model is highly unknown when applied to these varied and complex multi-agent structures.

In this paper, we take a pioneering step toward uncertainty estimation for LLM-based multi-agent systems by introducing a novel UQ framework of \textbf{M}ulti-\textbf{A}gent \textbf{T}ensor \textbf{U}ncertainty. To address the challenge of multi-step reasoning, MATU moves beyond analyzing only the final text, instead representing each agent's entire reasoning trajectory as an embedding matrix. To address the challenge of inter-agent communication diversity, we aggregate multiple runs of the same query to capture variability in how agents interact and exchange information. To address the challenge of communication topology diversity, MATU organizes all collected trajectories and runs into a higher-order tensor, which is inherently generalizable across different communication structures. This three-dimensional tensor, which is composed of agents, reasoning steps, and sampling runs, provides a holistic and generalizable way to represent the system's behavior. We can then apply tensor decomposition to disentangle and quantify the distinct sources of uncertainty, offering a comprehensive reliability measure at both the response and system levels.

\begin{itemize}
    \item We provide the first systematic definition of uncertainty quantification for LLM-based multi-agent systems, identifying unique sources of uncertainty introduced by tool usage, multi-step reasoning, and inter-agent communication in MAS.
    \item We design \ours, a tensor decomposition-based framework that integrates multi-agent uncertainty signals at both response and run levels, enabling holistic uncertainty estimation of multi-agent systems.
    \item We conduct extensive experiments across diverse tasks with or without tool-usage and communication topologies, and further provide case analyses that illustrate how different dimensions of uncertainty interact, demonstrating the need for dealing with the new challenge of UQ in multi-agent systems.
\end{itemize}

\section{Related Work}

\textbf{LLM-based Agents} LLMs have evolved into agents capable of solving diverse tasks, including web search~\citep{nakano2021webgpt,deng2023mind2web}, software development~\citep{wang2021codet5,yang2024swe}, and complex reasoning~\citep{gao2023pal,chen2022program}, by leveraging tools and historical memory~\citep{yao2023react, park2023generative,zhang2026selaur}. While single agents are effective, multi-agent systems (MAS) demonstrate superior performance through collaboration~\citep{li2023camel,wu2024autogen}. These systems employ varied communication topologies, ranging from static designs~\citep{li2023camel,qian2023communicative,hong2023metagpt,holt2023l2mac,zhou2023large}, prompt optimization~\cite{yao2026langmarl} to dynamic structures~\citep{zhuge2024gptswarm,liu2023dynamic,zhang2024g,wang2025anymac}. However, the complexity of these interactions poses new challenges for trustworthiness~
\cite{kirchhofposition}, necessitating uncertainty estimation methods that generalize across diverse agent topologies.

\vspace{1mm}
\noindent\textbf{Uncertainty for Large Language Model}
While uncertainty quantification is established for traditional regression and classification~\citep{ye2024uncertainty,amini2020deep,sensoy2018evidential,ovadia2019can}, LLMs' open-ended generation requires distinct approaches. Semantic entropy~\citep{kuhn2023semantic} addresses this but necessitates access to token probabilities. For black-box settings, recent works estimate uncertainty by analyzing the semantic consistency of generated responses~\citep{lin2023generating,chen2024quantifying,da2024llm,gao-etal-2024-spuq,hou2024decomposing}. These methods typically leverage NLI models to construct similarity matrices and derive uncertainty metrics from graph Laplacian eigenvalues~\citep{lin2023generating,chen2024quantifying,da2024llm,catak2024uncertainty}, or by integrating multiple uncertainty sources~\citep{chen2025uncertainty}.

However, research on UQ for agent systems remains underexplored, with \citet{kirchhof2025position,oh2026uncertainty} identifying key gaps in interactive and underspecification uncertainties. Currently, SAUP~\citep{zhao2025uncertainty} stands as the primary approach, employing situational weights for step-wise analysis. However, it treats steps independently, overlooking the holistic uncertainty of complete reasoning trajectories. In this paper, our work bridges this gap by integrating both response-level and run-level dynamics for a more reliable estimation of uncertainty
\section{Background}
\label{sec:background}

Uncertainty Quantification (UQ) for LLM-based agents extends beyond single-agent settings. 
In a multi-agent system (MAS), a set of $K$ agents $\{\mathcal{M}_1, \dots, \mathcal{M}_K\}$ 
collaboratively generate trajectories through communication and multi-step reasoning.  For different agents, the model parameter $\theta_k$ could be the same or different according to different designs. For different collaboration styles, the input of agent $M_i$ might also be different. For example, in a roundabout communication topology, the input might be the discussion contexts from other agents, while the input might be the assignment in a star communication topology. 

Considering previous UQ works on LLMs, repeated generations are key for the black-box UQ. Therefore, here we also define the repeated generations for MAS $S$. Given an input $x$, the $j$-th run of the MAS produces a trajectory $\tau^{(j)} = \{y^{(j,k)}_{1:T_k}\}_{k=1}^K$,
where $y^{(j,k)}_{1:T_k}$ denotes the sequence of outputs from agent $k$ during run $j$, and $T_k$ is the number of steps taken by agent $k$. For one task $x$, everything will be fixed, including the role and parameters of agents and communication topology. Then, across $N$ repeated runs of the MAS, we collect a set of trajectories 
$\mathcal{T} = \{\tau^{(1)}, \tau^{(2)}, \dots, \tau^{(N)}\}$.

\begin{problem}[Multi-agent Uncertainty]
Given an input $x$ and a set of trajectories $\mathcal{T}$ generated by a MAS across $N$ runs, 
the goal is to compute an uncertainty score $U$ that reflects the variability across $\mathcal{T}$. 
Formally,
$
U = \mathcal{F}(x, \mathcal{T}),
$
where $\mathcal{F}$ is an aggregation functional that maps the input and the trajectory set 
to a scalar value measuring the overall uncertainty. A lower $U$ indicates that the MAS 
consistently produces stable and reliable trajectories, while a higher $U$ suggests divergent 
reasoning, unstable communication, or fragile collaboration among agents. 
\end{problem}

Note that in our definition, each trajectory $\tau^{(j)}$ consists of agent-specific sequences $y^{(j,k)}_{1:T_k}$, where the horizon length $T_k$ may differ across agents and runs. This variability captures the intrinsic challenges of multi-step reasoning, since errors made in earlier steps can propagate differently depending on the trajectory length, and it also reflects the diversity of communication topologies, where agents may follow different interaction patterns.

\section{Method}
\begin{figure*}[h]
    \centering
    \includegraphics[width=\textwidth]{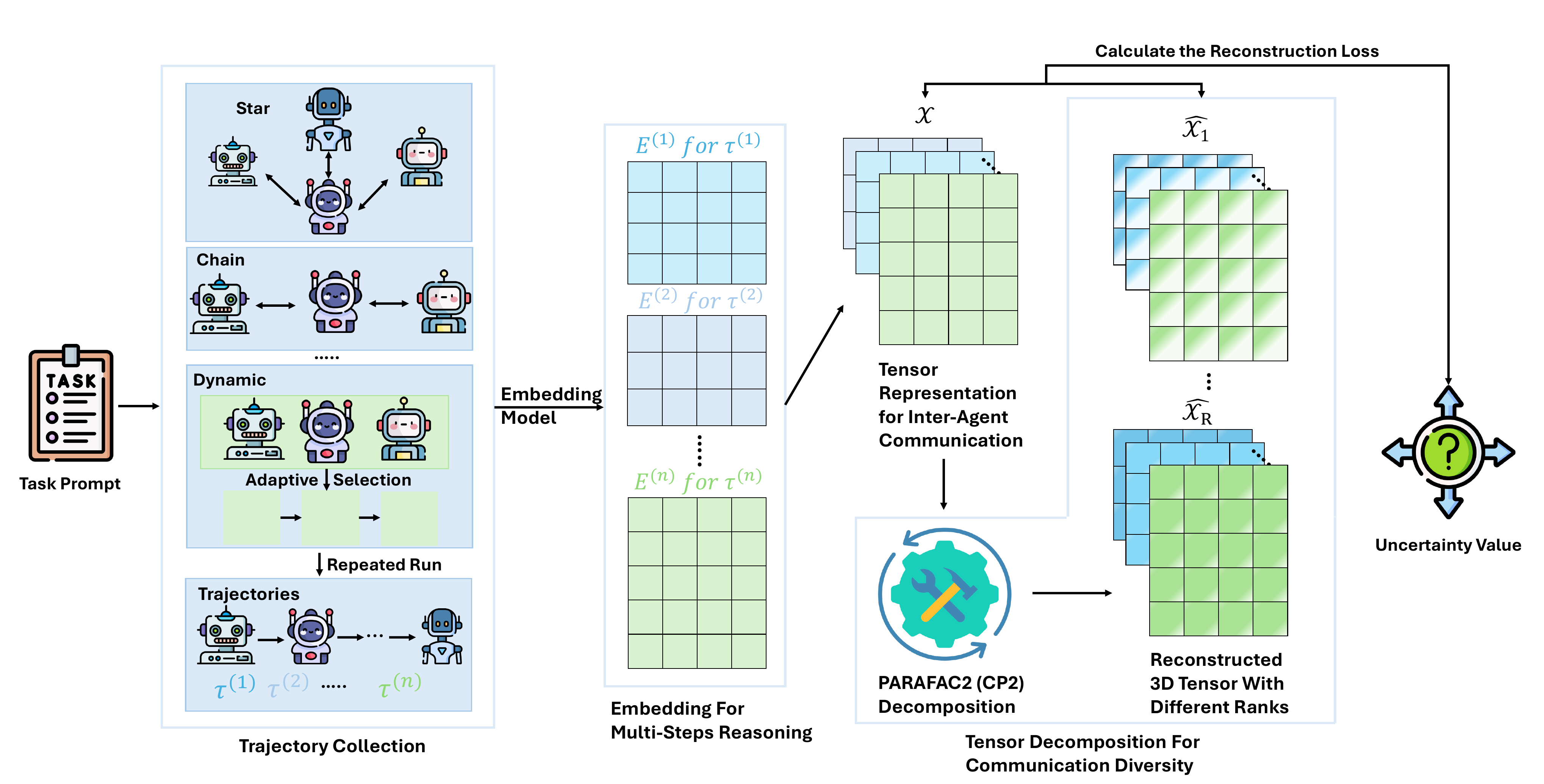}
    \caption{The overall pipeline of \ours. As shown in the figure, \ours could be applied to multi-agent systems with different communication topologies. We first collect trajectories for a fixed system and task, and then obtain embedding matrices for each trajectory. Then, we form a ragged tensor by stacking all embedding matrices and obtain the reconstructed tensor by conducting CP-2 decomposition. Finally, we use the reconstruction losses from reconstructed tensors with different ranks as the final uncertainty.}
    \label{fig:pipeline}
\end{figure*}

In this section, we present \ours in detail. We conceptualize multi-agent interactions as a \textbf{ragged tensor} to handle variable-length reasoning trajectories. The core logic is that consistent interactions across runs should follow a low-rank structure. By using tensor decomposition to find this low-rank approximation, the \textbf{reconstruction error} directly measures how much individual trajectories deviate from the shared consensus. A high error indicates that the runs are inconsistent, indicating a higher uncertainty. Therefore, we could directly use the reconstruction error as the uncertainty metric. A more theoretical explanation can be found at \cref{sec:appendix_theory} The overall pipeline of \ours is illustrated in \cref{fig:pipeline}, starting with the embedding process.

\subsection{Embedding for Multi-step Reasoning}

Multi-step reasoning poses a fundamental challenge in uncertainty quantification for multi-agent systems. Unlike single-turn settings, where the model outputs a single sentence, multi-agent reasoning unfolds as a sequence of intermediate steps. Errors 
introduced in earlier steps can cascade through subsequent ones, while different agents may take trajectories of varying lengths depending on their roles or communication topologies. This variability makes it difficult to directly compare 
trajectories across repeated runs.

To overcome these challenges, we encode each intermediate output, whether a natural language sentence or a tool call result, into a shared latent space using pre-trained embedding models. In detail, we treat tool call results as a string as well and use a text embedding model such as Qwen3-Embedding-0.6B.  Formally, for the $t$-th step in trajectory $\tau^{(j)}$, we define $e^{(j)}_t \in \mathbb{R}^d$,
where $d$ is the embedding dimension. By concatenating the embeddings across all steps in trajectory $j$ and agent $k$, we construct an embedding matrix $E^{(j,j)} \in \mathbb{R}^{T_{j,k} \times d}$,
where $T_{j,k}$ denotes the number of steps in that trajectory and agent. This embedding construction mitigates the core difficulties of multi-step reasoning by mapping heterogeneous, variable-length, and modality-mixed outputs to a fixed-dimensional semantic space at the step level, thereby decoupling semantic comparability from surface form and length. Semantically similar steps, even when expressed with different wording or produced by different agents or tools, are brought closer in the embedding space.  Besides, using additional embedding models facilitates step-wise aggregation without requiring token-level probabilities and establishes the foundation for subsequent tensor representations and decomposition.

\subsection{Tensor Representation for Inter-agent Communication}
The second challenge comes from inter-agent communication. Even when the agent system and the task input are fixed, MAS may produce distinct communication patterns. Agents can exchange information in slightly different orders, generate intermediate responses of different lengths, or invoke tools at different points. 

To capture such variability, we represent embedding matrices from repeated trajectories as a ragged tensor~\cite{fegade2022cora}. In run $j\in\{1,\ldots,N\}$, each agent $k\in\{1,\ldots,K\}$ produces a trajectory of length $T_{j,k}$, which we embed into a matrix
$E^{(j,k)} \in \mathbb{R}^{T_{j,k}\times d}$. We define the ragged object as the doubly-indexed matrix collection
\[
\mathcal{X}=\{\,E^{(j,k)} \mid j=1,\ldots,N;\; k=1,\ldots,K\,\},
\]
where $E^{(j,k)}$ denotes the stacked embedding matrix of agent $k$ in run $j$. Note that this matrix collection is a three-dimensional ragged tensor. Unlike a standard tensor in $\mathbb{R}^{N \times T \times d}$ that assumes a fixed $T$, the ragged tensor $\mathcal{X}$ allows $T_j$ to vary across runs:



\[
E^{(j,k)}\in\mathbb{R}^{T_{j,k}\times d}, \qquad T_{j,k}\neq T_{j',k'} \ \text{in general}.
\]

This representation enables us to aggregate multi-run trajectories into a single mathematical object without discarding the diversity of communication patterns. The variability of inter-agent communication is thus encoded directly into the 
structure of $\mathcal{X}$, laying the groundwork for decomposition methods that disentangle and quantify the uncertainty it induces.

\subsection{Tensor Decomposition for Communication Diversity}

The third challenge arises from communication diversity across different system topologies. Multi-agent systems may be organized in star~\citep{wu2024autogen}, chain~\citep{li2023camel}, or dynamic communication structures~\citep{wang2025anymac}, and each topology induces distinct statistical properties in the trajectories it generates. An uncertainty quantification framework must therefore be general enough to handle arbitrary topologies while remaining sensitive to their structural differences. 

To address this, we apply the PARAFAC2 Decomposition for Ragged Tensors (CP-2), a factorization method specifically designed to handle irregular tensor structures~\citep{schenker2023parafac2,perros2017spartan}. Unlike classical tensor decomposition, CP-2 operates directly on ragged tensors by aligning latent factors across dimensions of varying lengths. This property makes CP-2 particularly well-suited to our settings with variable lengths.

Formally, CP-2 seeks a low-rank approximation of the ragged tensor $\mathcal{X}$ in the form

\[
\mathcal{X} \approx \sum_{r=1}^{R} \lambda_r \, u_r^{(1)} \otimes u_r^{(2)} \otimes u_r^{(3)},
\]

where $R$ is the target rank, $\lambda_r$ are scalar weights, and $u_r^{(1)}, u_r^{(2)}, u_r^{(3)}$ are latent factors that are defined so as to respect the irregular lengths in $\mathcal{X}$. Through this decomposition, CP-2 captures shared patterns across steps, agents, and runs, while preserving the diversity introduced by different communication topologies. 

To quantify uncertainty, we perform CP-2 decomposition under different ranks $R$. For each $R$, we reconstruct an approximation $\hat{\mathcal{X}}_{R}$ and compute the reconstruction loss
$\mathcal{L}_{R} = \| \mathcal{X} - \hat{\mathcal{X}}_{R} \|$.

The sequence of losses $\{\mathcal{L}_{R}\}$ reflects how compressible the set of trajectories is under low-rank factors. Higher losses indicate that trajectories cannot be explained by a small number of latent components, implying higher uncertainty. To obtain a single scalar score, we aggregate the reconstruction losses across all considered ranks, defining the final uncertainty value as
\[
U = \sum_{R=1}^{R_{\max}} \mathcal{L}_{R},
\]

where $R_{\max}$ denotes the largest rank examined during decomposition. This score summarizes the degree to which variability in trajectories resists compression across different model capacities, and thus serves as the overall uncertainty estimate for the multi-agent system. By grounding the analysis in CP-2 decomposition, our framework can utilize the information from ragged tensors and generalize to arbitrary communication topologies while maintaining good uncertainty quantification.

\section{Experiments}
We conduct comprehensive experiments to evaluate the effectiveness of \ours. Our study is designed to answer the following research questions:\\
\noindent $\bullet$\textbf{RQ1:} Does \ours provide more accurate uncertainty quantification for multi-agent systems with static design?\\
\noindent $\bullet$ \textbf{RQ2:} Does \ours provide more accurate uncertainty quantification for multi-agent systems with dynamic design?\\
\noindent $\bullet$ \textbf{RQ3:} Does \ours provide more accurate uncertainty quantification for multi-agent systems with tool integration?

Beyond the research questions, we also provide a detailed case study to show why \ours could work in \cref{sec:case_study}.

\subsection{Experimental Setup}

\begin{table*}[h]
  \centering
  \begin{minipage}[t]{0.49\textwidth}
    \vspace{0pt}
    \centering
    \captionsetup{type=table}
    \caption{Comparison of our methods with different baselines on various datasets and LLMs on Camel~\cite{li2023camel}, \textbf{highlighted} with best performance.}
    \label{tab:main_results_camel}
    \vspace{-3mm}
    \setlength{\tabcolsep}{4pt}
    \renewcommand{\arraystretch}{1.1}
    \resizebox{\linewidth}{!}{%
    \begin{tabular}{lcccccc}
      \toprule
      \multirow{2}{*}{Methods} & \multicolumn{2}{c}{GPT-4o} & \multicolumn{2}{c}{Qwen2.5-7B} & \multicolumn{2}{c}{Llama3.1-8B} \\
      \cmidrule(lr){2-3} \cmidrule(lr){4-5} \cmidrule(lr){6-7}
      & AUROC & AUARC & AUROC & AUARC & AUROC & AUARC \\
      \midrule
      \multicolumn{7}{c}{\textbf{Dataset: MATH}} \\
      \midrule
      Eigv(Agre)-final     & 0.5698 & 0.5216 & 0.5238 &  0.8466 & 0.5243 & 0.6170 \\
      Eigv(Agre)-Whole   & 0.5632 & 0.5218  & 0.6784 & 0.8963 & 0.5622 & 0.6346 \\ 
      P(true)        & 0.5825 & 0.5592 & 0.6351 & 0.8855 & 0.5421 & 0.6303  \\ 
      SAUP-Single & - & - & 0.5597 & 0.8499 & 0.5244 & 0.6374\\
      SAUP-Multiple & - & - & 0.6078 & 0.8722 & 0.5258  & 0.6427\\
      \ours      & \textbf{0.6797} & \textbf{0.6160} & \textbf{0.7089}  & \textbf{0.9064} & \textbf{0.7354} & \textbf{0.7525} \\
      \midrule
      \multicolumn{7}{c}{\textbf{Dataset: MoreHopQA}} \\
      \midrule
      Eigv(Agre)-final     & 0.5307 & 0.3374 & 0.5631 & 0.6529 & 0.5572 & 0.5644 \\
      Eigv(Agre)-Whole   & 0.5259 & 0.3319 & 0.5420 & 0.6342 & 0.5398 & 0.5585 \\ 
      P(true)        & 0.5480 & 0.3405 & 0.5766 & 0.6512 & 0.5313 &  0.5460 \\ 
      SAUP-Single & - & - & 0.5103 & 0.6211 & 0.5083 & 0.5576  \\
      SAUP-Multiple & - & - & 0.5386 & 0.6345 & 0.5668 & 0.5798 \\
      \ours      & \textbf{0.5555} & \textbf{0.3474}  & \textbf{0.6529} & \textbf{0.7226} & \textbf{0.6320}  &  \textbf{0.6561} \\
      \midrule
      \multicolumn{7}{c}{\textbf{Dataset: MMLU}} \\
      \midrule
      Eigv(Agre)-final     & 0.5365 & 0.3304 & 0.5537 & 0.8023 & 0.5161  & 0.7270 \\
      Eigv(Agre)-Whole   & 0.5341 & 0.3236 & 0.5420 & 0.7995 & 0.5940 & 0.7646 \\ 
      P(true)        &  0.5059 & 0.3183 & 0.6846 & 0.8585 & 0.6207 & 0.7964 \\ 
      SAUP-Single &  -& - & 0.5233 & 0.7749 & 0.5424 & 0.7361 \\
      SAUP-Multiple & - & - & 0.5641 & 0.8100 & 0.5289 & 0.7330 \\
      \ours      & \textbf{0.5604} & \textbf{0.3384} & \textbf{0.7149} & \textbf{0.8656} & \textbf{0.7075} & \textbf{0.8427} \\
      \bottomrule
    \end{tabular}}
  \end{minipage}
  \hfill
  \begin{minipage}[t]{0.49\textwidth}
    \vspace{0pt}
    \centering
    \captionsetup{type=table}
    \caption{Comparison of our methods with different baselines on various datasets and LLMs on AutoGen~\cite{wu2024autogen}, \textbf{highlighted} with best performance}
    \label{tab:main_results_autogen}
    \vspace{-3mm}
    \setlength{\tabcolsep}{4pt}
    \renewcommand{\arraystretch}{1.1}
    \resizebox{\linewidth}{!}{%
    \begin{tabular}{lcccccc}
      \toprule
      \multirow{2}{*}{Methods} & \multicolumn{2}{c}{GPT-4o} & \multicolumn{2}{c}{Qwen2.5-7B} & \multicolumn{2}{c}{Llama3.1-8B} \\
      \cmidrule(lr){2-3} \cmidrule(lr){4-5} \cmidrule(lr){6-7}
      & AUROC & AUARC & AUROC & AUARC & AUROC & AUARC \\
      \midrule
      \multicolumn{7}{c}{\textbf{Dataset: MATH}} \\
      \midrule
      Eigv(Agre)-final     & 0.5898 & 0.5826 & 0.6355 & 0.4512 & 0.5912  & 0.3802 \\
      Eigv(Agre)-Whole   & 0.6015 & 0.5892 & 0.6111 & 0.4326 & 0.5761 & 0.3679 \\ 
      P(true)        & 0.6079 & 0.5931 & 0.6524 & 0.5102 & 0.6271 & 0.4571 \\ 
      SAUP-Single & - & - & 0.5268 &  0.3990 &  0.6064 &  0.3830\\
      SAUP-Multiple & - & - & 0.5385  & 0.4090 &  0.6334 & 0.3933 \\
      \ours      & \textbf{0.6582} & \textbf{0.6220} & \textbf{0.7146} & \textbf{0.5334} & \textbf{0.7544} & \textbf{0.4687} \\
      \midrule
      \multicolumn{7}{c}{\textbf{Dataset: MoreHopQA}} \\
      \midrule
      Eigv(Agre)-final     & 0.5311 & 0.4968 & 0.5331 & 0.6678 & 0.5395 & 0.5721 \\
      Eigv(Agre)-Whole   & 0.5218 & 0.4942 & 0.5323 & 0.6689 & 0.5279 & 0.5642 \\ 
      P(true)        & 0.5598 & 0.5033 & 0.5806 & 0.7031 & 0.5515 & 0.5827 \\ 
      SAUP-Single & - & - &  0.5197 & 0.6445 & 0.5422 & 0.5782 \\
      SAUP-Multiple & - & - &  0.5342 & 0.6708 & 0.5488 & 0.5877 \\
      \ours      & \textbf{0.5817} & \textbf{0.5237} & \textbf{0.6392} & \textbf{0.7374} & \textbf{0.5989} & \textbf{0.6117} \\
      \midrule
      \multicolumn{7}{c}{\textbf{Dataset:MMLU}} \\
      \midrule
      Eigv(Agre)-final     & 0.5981 & 0.5649 & 0.7105 & 0.8617 & 0.5521 & 0.4288 \\
      Eigv(Agre)-Whole   & 0.5759 & 0.5438 & 0.6867 & 0.8516 & 0.5316 & 0.3762 \\ 
      P(true)        & 0.5802 & 0.5528 & 0.6556  & 0.8363 &0.5775  & 0.4368 \\ 
      SAUP-Single & - & - & 0.6484 & 0.8552 & 0.5138  & 0.3031\\
      SAUP-Multiple & - & - & 0.7193 & 0.8589 & 0.5018 & 0.2973 \\
      \ours      & \textbf{0.6277} & \textbf{0.5841} & \textbf{0.7315} & \textbf{0.8833}  &  \textbf{0.5954}   &  \textbf{0.4745} \\
      \bottomrule
    \end{tabular}}
  \end{minipage}
\end{table*}

\textbf{Dataset} To comprehensively evaluate \ours, we use four diverse datasets: MATH (mathematical reasoning)~\cite{hendrycks2021measuring}, MoreHopQA (multi-hop QA)~\cite{Schnitzler2024MoreHopQAMT}, MMLU (general knowledge)~\citep{hendrycks2020measuring}, and HumanEval (code generation)~\citep{chen2021evaluating}. Detailed descriptions are provided in \cref{sec:appendix_experimental_setting}.


\noindent \textbf{Multi-agent System}. We use multiple MAS with different designs. In detail, we consider using Camel~\citep{li2023camel}, which consists of an AI User and an AI Assistant with round-robin conversation, and AutoGen~\citep{wu2024autogen}, which uses a star agent that assigns tasks to all other agents. Both frameworks use static design. On the other hand, we use AnyMac~\citep{wang2025anymac}, which will dynamically choose the next agent based on the progress as the dynamic multi-agent system.

\noindent \textbf{Models} For models behind agents, we are using both open-source and closed-source models. For the open-source model, we mainly use Qwen2.5-7B~\cite{bai2023qwen} and Llama3.1-8B~\cite{dubey2024llama}, which is the representative open-source model. For closed-source models, we mainly use GPT-4o from OpenAI.

\noindent \textbf{Evaluation Metrics} Effective uncertainty measures should correlate with response correctness: higher uncertainty should indicate a higher likelihood of error. Following prior work~\cite{lin2023generating,da2024llm}, we evaluate uncertainty estimates by using them to predict whether a generated answer is correct. We report Area Under Receiver Operating Characteristic (AUROC) and Area Under Accuracy Rejection Curve (AUARC) as evaluation metrics, where \textbf{a higher AUROC or AUARC demonstrates better uncertainty measures}. To compute AUROC and AUARC, the accuracy of each original response is required. To label responses as correct or incorrect, we use a reference LLM, GPT-5, to provide correctness scores to the final answer from MAS. 

\noindent \textbf{Baseline} We compare MATU against three baselines: P(true)~\citep{kadavath2022language}, Eigv(Agr)~\citep{lin2023generating}, and SAUP~\citep{zhao2024saup}. For the Eigv(Agr), we use the final answer or every conversation to compute the entailment matrix~\cite{bowman2015large}, resulting in two different variants: Eigv(Agr)-Answer and Eigv(Agr)-Whole. SAUP is originally designed for one trajectory, while we collect multiple trajectories. Therefore, we use SAUP-Single, which uses the SAUP from the first trajectory, and SAUP-Multiple which uses the mean SAUP from all trajectories. Please note that SAUP is a \textbf{white-box} method so that it cannot be applied to closed-source models. More introduction can be found at \cref{sec:appendix_baseline}.

\noindent \textbf{Implementation Detail} For the embedding models, we use off-the-shelf Qwen3-embedding-0.6B to get the fast processing speed. For trajectories, we collect 10 trajectories for every task, and we use a temperature of 0.9 for every setting. All the experiments are conducted on a single Nvidia A100-80GB GPU or using an OpenAI API.

\subsection{Performance for Multi-agent System with Static Design (RQ1)}
Firstly, to explore how good \ours is for MAS with static design, we conduct experiments on Camel~\cite{li2023camel} and AutoGen~\cite{wu2024autogen} and three different datasets to demonstrate the performance comprehensively. The results are shown in \cref{tab:main_results_camel} and \cref{tab:main_results_autogen}. The results show that:

\noindent  $\bullet$ \ours consistently outperforms all baselines by capturing holistic system-level behavior rather than just final output consistency. While traditional methods like Eigv(Agre) focus on semantic similarity and SAUP measures step-wise uncertainty independently, \ours integrates the entire reasoning trajectory and multi-run communication patterns into a unified tensor. This approach allows it to identify fragile consensus in the collaborative process that response-level or single-trajectory measures fail to detect.

\noindent $\bullet$ \ours shows consistent reliability whether the task involves challenging mathematical reasoning in MATH, multi-hop question-answering in MorehopQA, or broad knowledge synthesis in MMLU. By mapping heterogeneous outputs into a shared embedding space, the method provides a robust reliability measure that remains effective regardless of whether the MAS is performing logical deduction or knowledge retrieval.

\begin{table}[t]
  \centering
  \caption{Comparison of our methods with different baselines on various datasets and LLMs on AnyMac~\cite{wang2025anymac}, \textbf{highlighted} with best performance.}
  \label{tab:main_results_AnyMac}
    \vspace{-3mm}
  \setlength{\tabcolsep}{4pt}
  \renewcommand{\arraystretch}{1.1}
  \resizebox{\linewidth}{!}{%
  \begin{tabular}{lcccccc}
    \toprule
    \multirow{2}{*}{Methods} & \multicolumn{2}{c}{GPT-4o} & \multicolumn{2}{c}{Qwen2.5-7B} & \multicolumn{2}{c}{Llama3.1-8B} \\
    \cmidrule(lr){2-3} \cmidrule(lr){4-5} \cmidrule(lr){6-7}
    & AUROC & AUARC & AUROC & AUARC & AUROC & AUARC \\
    \midrule
    \multicolumn{7}{c}{\textbf{Dataset: MATH}} \\
    \midrule
    Eigv(Agre)-final     & 0.6359 & 0.6133 & 0.6506 & 0.8212 & 0.6340 & 0.6059 \\
    Eigv(Agre)-Whole   & 0.6308 & 0.6115 & 0.6314 & 0.8081 & 0.6215  & 0.5953 \\ 
    P(true)        & 0.6226 & 0.6070 & 0.6602 & 0.8291 & 0.6581 & 0.6225  \\ 
    SAUP-Single & - & - & 0.6261 & 0.7982 & 0.6339 & 0.6008 \\
    SAUP-Multiple & - & - & 0.6396 & 0.8119 & 0.6477 & 0.6065 \\
    MATU      & \textbf{0.6675} & \textbf{0.6439} & \textbf{0.6966} & \textbf{0.8585} & \textbf{0.7121} & \textbf{0.6518} \\
    \midrule
    \multicolumn{7}{c}{\textbf{Dataset: MorehopQA}} \\
    \midrule
    Eigv(Agre)-final     & 0.5257 & 0.3992 & 0.6079 & 0.6741 & 0.6110 & 0.6369 \\
    Eigv(Agre)-Whole   & 0.5203 & 0.4010 & 0.6021 & 0.6681 & 0.6034 & 0.6300 \\ 
    P(true)        & 0.5455 & 0.4121 & 0.6205 & 0.6853 & 0.6158 & 0.6416 \\ 
    SAUP-Single & - & - & 0.6088 & 0.6770 & 0.5918 & 0.6277 \\
    SAUP-Multiple & - & - & 0.6242 & 0.6914 & 0.6055 & 0.6322 \\
    MATU      & \textbf{0.5671} & \textbf{0.4336} & \textbf{0.6457} & \textbf{0.7029} & \textbf{0.6262} & \textbf{0.6493}  \\
    \midrule
    \multicolumn{7}{c}{\textbf{Dataset: MMLU}} \\
    \midrule
    Eigv(Agre)-final     & 0.5568 & 0.4952 & 0.5446 & 0.7650  & 0.5321 & 0.6586 \\
    Eigv(Agre)-Whole   & 0.5641 & 0.5049 & 0.5337 & 0.7433 & 0.5215 & 0.6512 \\ 
    P(true)        & 0.5594 & 0.4976 & 0.5552 & 0.7681 & 0.5297 & 0.6632 \\ 
    SAUP-Single & - & - & 0.5048 & 0.7261 & 0.5340  & 0.6542 \\
    SAUP-Multiple & - & - & 0.5382 & 0.7602 & 0.5382 & 0.6719 \\
    MATU      & \textbf{0.5925} & \textbf{0.5152} & \textbf{0.5821} & \textbf{0.7768} &  \textbf{0.5500} & \textbf{0.6797}  \\
    \bottomrule
  \end{tabular}}
\end{table}

\subsection{Performance for Multi-agent System with Dynamic Design (RQ2)}

To evaluate the performance of \ours in more complex, adaptive environments, we extend our evaluation to multi-agent systems with dynamic designs. Unlike static topologies, dynamic systems such as AnyMac~\cite{wang2025anymac} adaptively select the next agent during execution based on the evolving context of the task. We conduct these experiments across the same datasets using both open-source models and closed-source architectures (GPT-4o). The results for the AnyMac system are detailed in \cref{tab:main_results_AnyMac}, leading to the following observations:

\noindent $\bullet$ \ours demonstrates superior adaptability to unpredictable communication sequences by leveraging higher-order tensor representations. In dynamic systems where the sequence of agent interactions varies significantly between runs, traditional semantic or step-wise baselines struggle to maintain a consistent reliability measure. By organizing these varied trajectories into a ragged tensor and applying tensor decomposition, \ours successfully aligns latent factors across dimensions of varying lengths, allowing it to outperform the strongest baselines by a significant margin in AUROC and AUARC.

\begin{figure*}[htbp]
    \centering
    \begin{minipage}{0.31\textwidth}
        \centering
        \includegraphics[width=\linewidth]{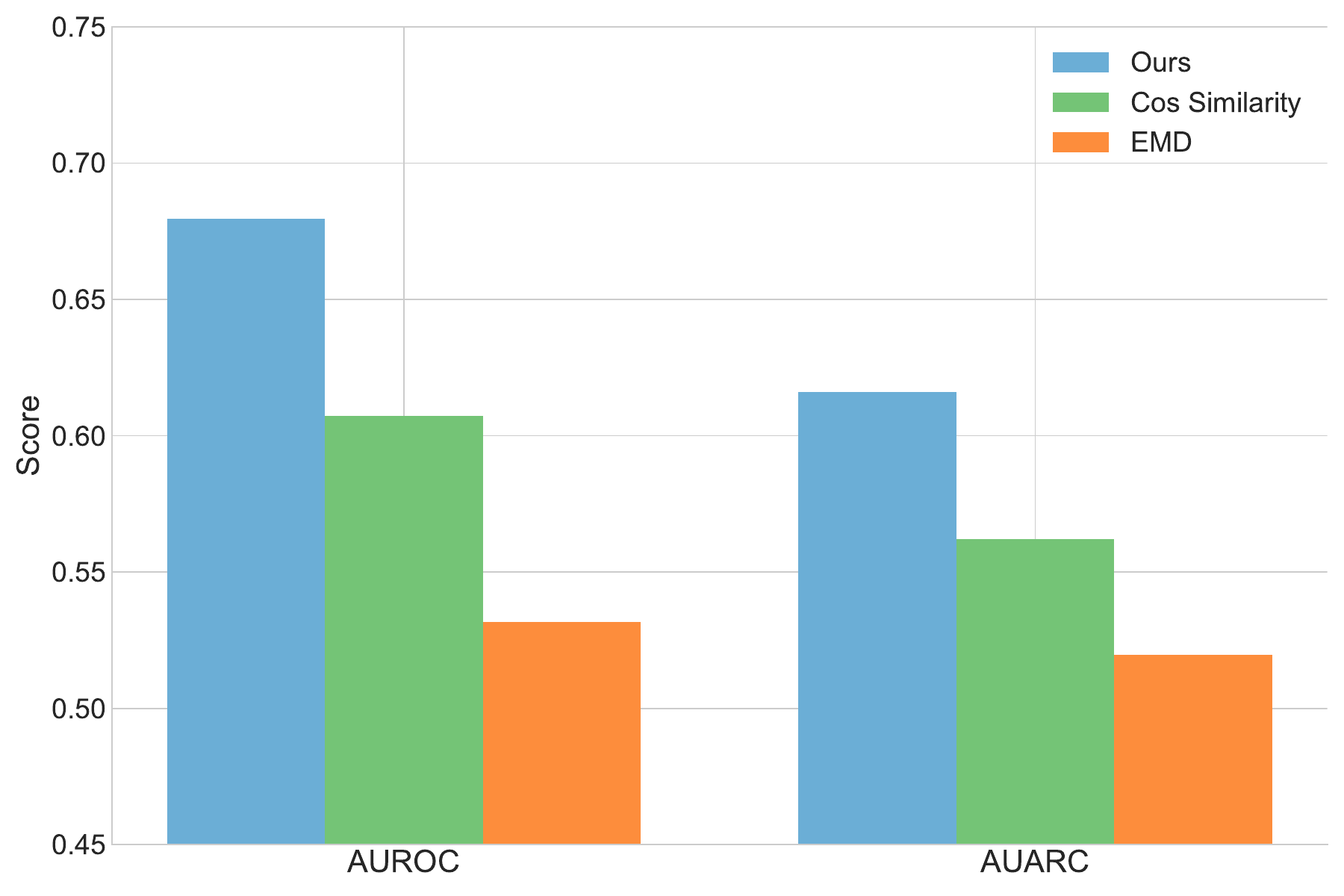}
        \subcaption{Ablation Study}
        \label{fig:variant}
    \end{minipage}
    \hfill
    \begin{minipage}{0.31\textwidth}
        \centering
        \includegraphics[width=\linewidth]{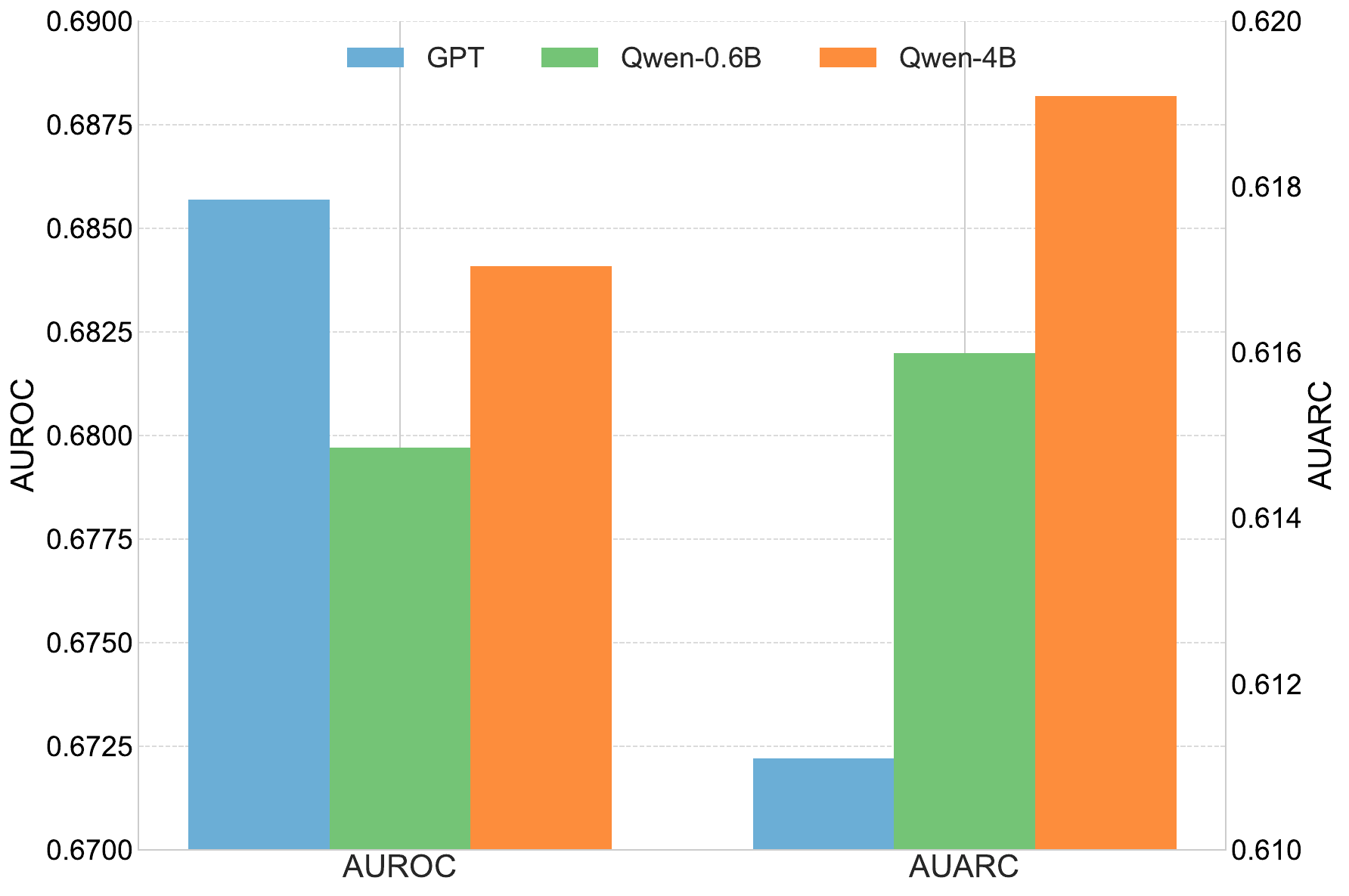}
        \subcaption{Embedding Models}
        \label{fig:embedding}
    \end{minipage}
    \hfill
    \begin{minipage}{0.31\textwidth}
        \centering
        \includegraphics[width=\linewidth]{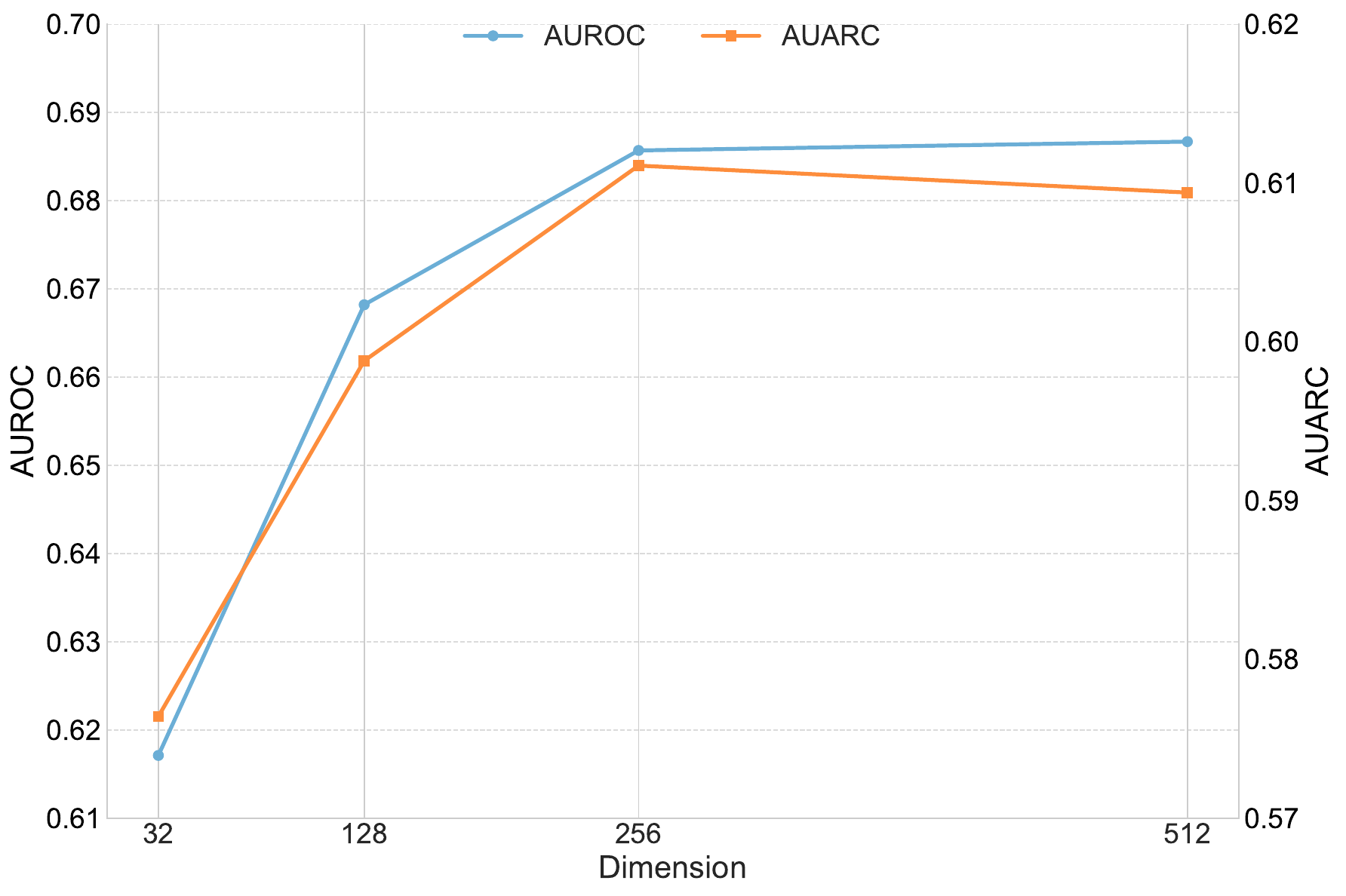}
        \subcaption{Effect of Embedding Dimension}
        \label{fig:dimension}
    \end{minipage}
    \vspace{-2mm}
    \caption{Results for ablation study and sensitivity study. The results show that our design for \ours and our choices of the hyperparameter are well-suited.}
    \label{fig:ablation}
\end{figure*}

\noindent  $\bullet$ Comparative analysis of self-evaluation and propagation baselines underscores the necessity of multi-run structural ensembling. Considering all experimental results and the baselines, $P(true)$ emerges as the most competitive, likely because it leverages the LLM's intrinsic ability to reflect on its own non-linear reasoning process. Furthermore, the consistent superiority of SAUP\-Multiple over SAUP\-Single confirms that a single interaction path is a poor proxy for the system's overall reliability in a dynamic environment.

\subsection{Performance for Multi-agent System with Tool Integration (RQ3)}

\begin{figure}[h]
    \centering
    \includegraphics[width=0.45\textwidth]{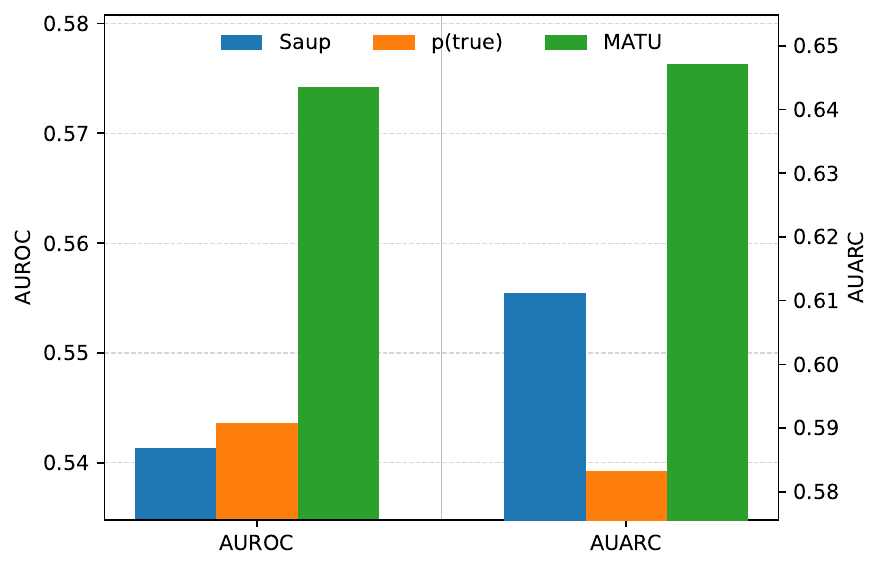}
    \vspace{-3mm}
    \caption{Comparison of \ours and baselines on llama3 and the Humaneval dataset. The results show that \ours can have better results even with tool integration, showing the robustness of \ours.}
    \label{fig:humaneval}
\end{figure}

To explore whether \ours can effectively quantify the reliability of collaborative agents when integrated with external tools, we conduct experiments on the HumanEval benchmark. This task requires agents not only to reason linguistically but also to synthesize executable code and interact with a Python interpreter, which serves as a functional tool within the multi-agent workflow. We conduct the experiments on llama3, and the results can be found at \cref{fig:humaneval}. The results show that \ours outperforms other baselines on both AUROC and AUARC with the Humaneval dataset, showing the robustness of \ours with code environment and tool integration.

\subsection{Ablation and Sensitivity Study}

To further analyze the robustness and key components of our framework, we conduct a series of ablation and sensitivity experiments on the Camel and the MATH dataset with GPT-4o.

\noindent \textbf{Ablation with Input Variants} To verify whether raw embedding tensors are superior to traditional distance-based representations, we compare \ours against variants that use Earth Mover’s Distance (EMD) and Cosine Similarity to construct the similarity matrices for decomposition instead of the step-level embedding matrices. As shown in \cref{fig:variant}, \ours consistently yields higher AUROC and AUARC scores, while EMD and Cosine Similarity fail to capture the granular latent signals within agent trajectories. This confirms that applying tensor decomposition directly to reasoning embeddings preserves significantly richer multi-agent dynamics than distance-based metrics.

\noindent \textbf{Impact of Embedding Models} We examine how the choice of the underlying text embedding model affects the precision of uncertainty estimation. We evaluate three models of varying scales: GPT-Embedding, Qwen-0.6B-Embedding, and Qwen-4B-Embedding.  The results are shown in \cref{fig:embedding}. While larger models like Qwen-4B and GPT-Embedding provide slight performance gains, the difference compared to the Qwen-0.6B model is minimal. We conclude that Qwen-0.6B offers the optimal balance between computational efficiency and accuracy, making it sufficient for UQ.

\noindent  \textbf{Sensitivity to Embedding Dimensions} To determine the optimal latent space dimensionality for representing complex reasoning steps, we evaluate the system's performance across dimensions ranging from 32 to 512. The results in \cref{fig:embedding} indicate a sharp improvement in both AUROC and AUARC as the dimension increases to 256, after which the gains become marginal. Consequently, we select 256 as our default embedding dimension to ensure comprehensive representation without incurring redundant computational overhead.

\noindent \textbf{Semantic Interpretation of Latent Factors} Beyond scoring uncertainty, \ours provides interpretability by mapping PARAFAC2 latent components to specific reasoning failures. For example, in our MATH case study, a component capturing a semantic misinterpretation exhibited a $2.29\times$ higher mean factor loading in incorrect runs (1.15) than in correct ones (0.50). This demonstrates \ours's ability to localize recurring errors statistically. Detailed interpretation protocols and verbatim logs are provided in Appendix \ref{sec:appendix_interpret}.

\subsection{Performance under OOD Situation}

\begin{table}[t!]
\centering
\caption{Uncertainty quantification performance (AUROC) under strict distribution shift (Math prompts on MMLU dataset) using Qwen2.5-7B.}
\label{tab:ood}
    \vspace{-3mm}
\small
\begin{tabular}{lc}
\toprule
\textbf{Method} & \textbf{AUROC ($\uparrow$)} \\
\midrule
SAUP-Single & 0.5930 \\
SAUP-Multiple & 0.6201 \\
Eigv(Agre)-final & 0.6355 \\
\midrule
\textbf{\ours} & \textbf{0.6770} \\
\bottomrule
\end{tabular}
\end{table}

To evaluate the performance of our framework under strict out-of-domain (OOD) scenarios, we simulate a distribution shift by applying our highly specialized, math-specific multi-agent prompts to non-mathematical general knowledge tasks from the MMLU dataset (using Qwen2.5-7B). As shown in \cref{tab:ood}, \ours robustly captures system uncertainty in this misaligned setting, achieving an AUROC of 0.6770 and significantly outperforming established baselines. Furthermore, we investigate whether \ours can be utilized for OOD sample detection. Our empirical results show that the normalized uncertainty score from \ours exhibits a clear separation between in-distribution data (0.13, using MMLU prompts on MMLU) and OOD data (0.92, using Math prompts on MMLU). In contrast, the baseline SAUP struggles to distinguish between the two (0.29 vs. 0.46). This confirms that \ours not only generalizes well to unseen domains but also effectively serves as an OOD detector.

\subsection{Down-stream Task: Model Selection}

\begin{figure}[t]
    \centering
    \includegraphics[width=0.45\textwidth]{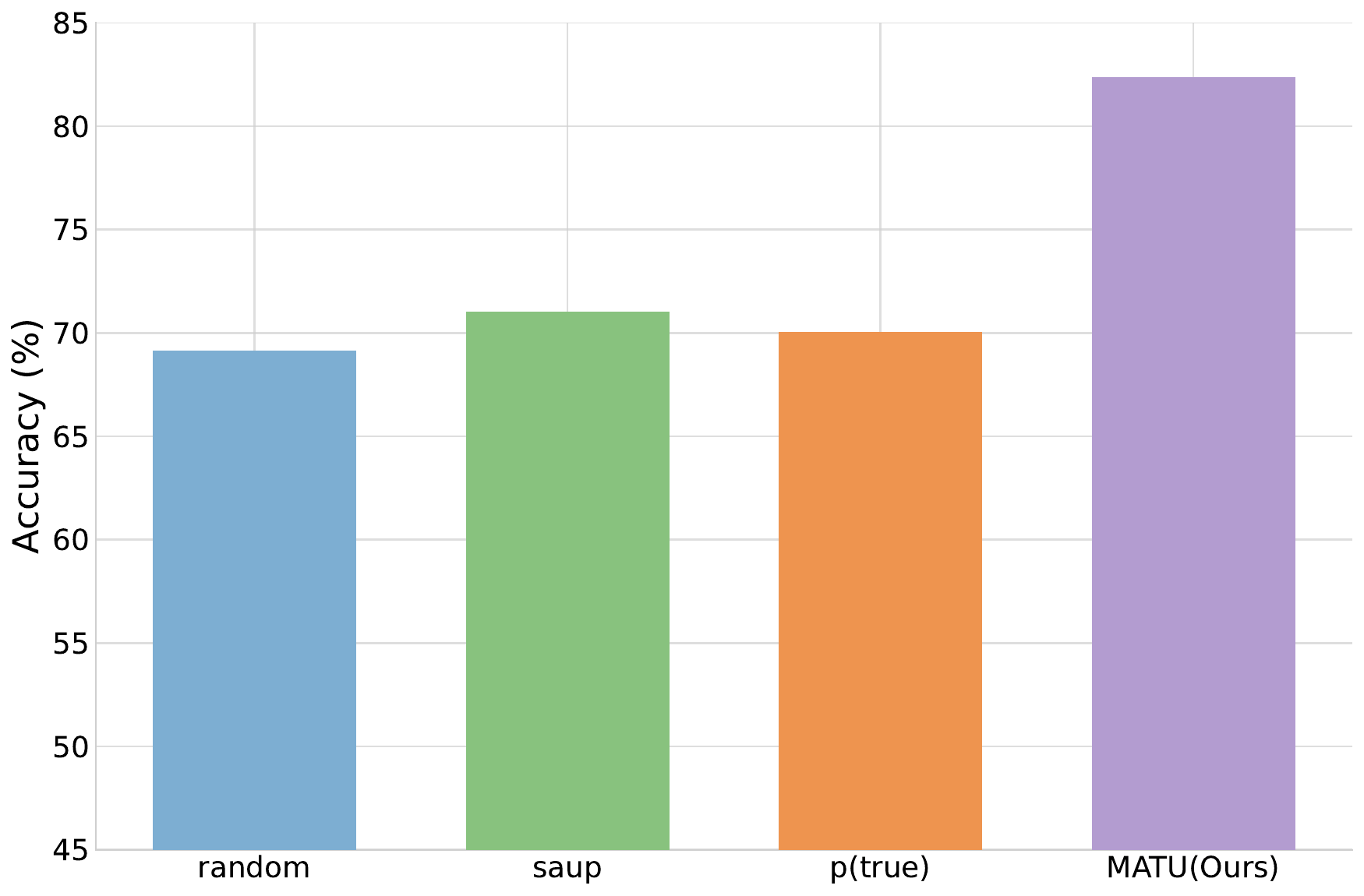}
    \vspace{-1mm}
    \caption{Comparison of backbone selection results. A higher accuracy demonstrates a better selection strategy. The results show that \ours has a superior performance improvement on accuracy, indicating that \ours offers a more robust uncertainty value.}
    \label{fig:Downstream}
\end{figure}

To evaluate the practical utility of $\ours$ in real-world deployment, we conduct a backbone selection task. This experiment explores whether uncertainty scores can serve as a reliable signal to select the most accurate answer from a pool of different MAS configurations. Specifically, for a given query, we generate multiple potential solutions across four distinct LLM backbones: Qwen2.5-7B~\cite{Yang2024Qwen25TR}, Llama3.1-8B~\cite{dubey2024llama}, Qwen3-4B~\cite{Yang2025Qwen3TR}, and Gemma3-4B~\cite{Kamath2025Gemma3T}. For each query, the system identifies the backbone that yields the lowest uncertainty score $U$ and selects its response as the final output.  We evaluate this routing strategy by comparing the resulting system accuracy when guided by $\ours$ against selection based on $P(true)$, $SAUP\text{-Multiple}$, and a random selection baseline on Camel framework~\cite{li2023camel} and the MATH dataset~\cite{hendrycks2021measuring}. The results are shown in \cref{fig:Downstream}. The results show the superior performance improvement using \ours, showing that \ours is a robust tool for backbone selection, which indicates that \ours offers a robust uncertainty value.

\section{Conclusion}

In this work, we propose \ours, a pioneering framework for quantifying uncertainty in LLM-based multi-agent systems by leveraging tensor decomposition to capture the holistic dynamics of multi-step reasoning and inter-agent communication. By organizing reasoning trajectories into ragged tensors and analyzing them via PARAFAC2 decomposition, our method effectively disentangles sources of uncertainty across varying communication topologies and run lengths, overcoming the limitations of traditional semantic or step-wise approaches. Extensive experiments on diverse benchmarks demonstrate that MATU consistently outperforms existing baselines in both static and dynamic system designs, while also proving its practicality in downstream tasks such as backbone model selection.

\section*{Acknowledgment}
The work was partially supported by NSF award \#2442477 and \#2550203. We thank Amazon Research Awards, Cisco Faculty Research Awards, and Toyota Faculty Research Awards. The authors acknowledge Google and OpenAI for providing us with API credits and Research Computing at Arizona State University for providing computing resources. 
The work at UCR was partially supported by the NSF under CAREER grant IIS \#2046086, grant \#2431569 and
CREST Center for Multidisciplinary Research Excellence in CyberPhysical Infrastructure Systems (MECIS) grant \#2112650
The views and conclusions in this paper are those of the authors and should not be interpreted as representing any funding agencies.

\section*{Limitations}

While MATU demonstrates effectiveness in quantifying uncertainty for multi-agent systems, we acknowledge several limitations in our current work. First, the core mechanism relies on constructing a higher-order tensor from multiple reasoning trajectories (e.g., $N=10$ runs in our experiments), meaning the inference cost scales linearly with the number of sampled trajectories. Although this multi-run paradigm is standard in black-box uncertainty estimation, it inevitably consumes more computational resources compared to single-pass methods. Second, since MATU decouples semantic meaning from surface form by mapping reasoning steps into a latent space, the sensitivity and accuracy of our uncertainty quantification are bounded by the quality of the underlying embedding model. In highly specialized domains where general-purpose embedding models may fail to capture subtle semantic nuances, MATU's performance might degrade unless domain-specific embeddings are employed. 

\bibliography{custom}

@article{da2024open,
  title={Open-ti: Open traffic intelligence with augmented language model},
  author={Da, Longchao and Liou, Kuanru and Chen, Tiejin and Zhou, Xuesong and Luo, Xiangyong and Yang, Yezhou and Wei, Hua},
  journal={International Journal of Machine Learning and Cybernetics},
  volume={15},
  number={10},
  pages={4761--4786},
  year={2024},
  publisher={Springer}
}

@article{zhang2026selaur,
  title={SELAUR: Self Evolving LLM Agent via Uncertainty-aware Rewards},
  author={Zhang, Dengjia and Liu, Xiaoou and Cheng, Lu and Wang, Yaqing and Murray, Kenton and Wei, Hua},
  journal={arXiv preprint arXiv:2602.21158},
  year={2026}
}

@inproceedings{kirchhofposition,
  title={Position: Uncertainty Quantification Needs Reassessment for Large Language Model Agents},
  author={Kirchhof, Michael and Kasneci, Gjergji and Kasneci, Enkelejda},
  booktitle={Forty-second International Conference on Machine Learning Position Paper Track},
  year={2025}
}

@inproceedings{yao2025comal,
  title={Comal: Collaborative multi-agent large language models for mixed-autonomy traffic},
  author={Yao, Huaiyuan and Da, Longchao and Nandam, Vishnu and Turnau, Justin and Liu, Zhiwei and Pang, Linsey and Wei, Hua},
  booktitle={Proceedings of the 2025 SIAM International Conference on Data Mining (SDM)},
  pages={409--418},
  year={2025},
  organization={SIAM}
}

@article{da2025understanding,
  title={Understanding the uncertainty of llm explanations: A perspective based on reasoning topology},
  author={Da, Longchao and Liu, Xiaoou and Dai, Jiaxin and Cheng, Lu and Wang, Yaqing and Wei, Hua},
  journal={arXiv preprint arXiv:2502.17026},
  year={2025}
}

@article{yao2026langmarl,
  title={LangMARL: Natural Language Multi-Agent Reinforcement Learning},
  author={Yao, Huaiyuan and Da, Longchao and Liu, Xiaoou and Fleming, Charles and Chen, Tianlong and Wei, Hua},
  journal={arXiv preprint arXiv:2604.00722},
  year={2026}
}

@article{amini2020deep,
  title={Deep evidential regression},
  author={Amini, Alexander and Schwarting, Wilko and Soleimany, Ava and Rus, Daniela},
  journal={Advances in neural information processing systems},
  volume={33},
  pages={14927--14937},
  year={2020}
}

@inproceedings{xia2025survey,
  title={A survey of uncertainty estimation methods on large language models},
  author={Xia, Zhiqiu and Xu, Jinxuan and Zhang, Yuqian and Liu, Hang},
  booktitle={Findings of the Association for Computational Linguistics: ACL 2025},
  pages={21381--21396},
  year={2025}
}

@article{oh2026uncertainty,
  title={Uncertainty Quantification in LLM Agents: Foundations, Emerging Challenges, and Opportunities},
  author={Oh, Changdae and Park, Seongheon and Kim, To Eun and Li, Jiatong and Li, Wendi and Yeh, Samuel and Du, Xuefeng and Hassani, Hamed and Bogdan, Paul and Song, Dawn and others},
  journal={arXiv preprint arXiv:2602.05073},
  year={2026}
}

@inproceedings{liu2025uncertainty,
  title={Uncertainty quantification and confidence calibration in large language models: A survey},
  author={Liu, Xiaoou and Chen, Tiejin and Da, Longchao and Chen, Chacha and Lin, Zhen and Wei, Hua},
  booktitle={Proceedings of the 31st ACM SIGKDD Conference on Knowledge Discovery and Data Mining V. 2},
  pages={6107--6117},
  year={2025}
}

@inproceedings{li2024urbangpt,
  title={Urbangpt: Spatio-temporal large language models},
  author={Li, Zhonghang and Xia, Lianghao and Tang, Jiabin and Xu, Yong and Shi, Lei and Xia, Long and Yin, Dawei and Huang, Chao},
  booktitle={Proceedings of the 30th ACM SIGKDD conference on knowledge discovery and data mining},
  pages={5351--5362},
  year={2024}
}

@inproceedings{yao-etal-2026-instructional,
    title = "Instructional Agents: Reducing Teaching Faculty Workload through Multi-Agent Instructional Design",
    author = "Yao, Huaiyuan  and
      Xu, Wanpeng  and
      Turnau, Justin  and
      Kellam, Nadia  and
      Wei, Hua",
    editor = "Demberg, Vera  and
      Inui, Kentaro  and
      Marquez, Llu{\'i}s",
    booktitle = "Proceedings of the 19th Conference of the {E}uropean Chapter of the {A}ssociation for {C}omputational {L}inguistics (Volume 1: Long Papers)",
    month = mar,
    year = "2026",
    address = "Rabat, Morocco",
    publisher = "Association for Computational Linguistics",
    url = "https://aclanthology.org/2026.eacl-long.191/",
    doi = "10.18653/v1/2026.eacl-long.191",
    pages = "4087--4109",
    ISBN = "979-8-89176-380-7"
}

@inproceedings{yao2023react,
  title={React: Synergizing reasoning and acting in language models},
  author={Yao, Shunyu and Zhao, Jeffrey and Yu, Dian and Du, Nan and Shafran, Izhak and Narasimhan, Karthik and Cao, Yuan},
  booktitle={International Conference on Learning Representations (ICLR)},
  year={2023}
}

@inproceedings{park2023generative,
  title={Generative agents: Interactive simulacra of human behavior},
  author={Park, Joon Sung and O'Brien, Joseph and Cai, Carrie Jun and Morris, Meredith Ringel and Liang, Percy and Bernstein, Michael S},
  booktitle={Proceedings of the 36th annual acm symposium on user interface software and technology},
  pages={1--22},
  year={2023}
}

@article{nakano2021webgpt,
  title={Webgpt: Browser-assisted question-answering with human feedback},
  author={Nakano, Reiichiro and Hilton, Jacob and Balaji, Suchir and Wu, Jeff and Ouyang, Long and Kim, Christina and Hesse, Christopher and Jain, Shantanu and Kosaraju, Vineet and Saunders, William and others},
  journal={arXiv preprint arXiv:2112.09332},
  year={2021}
}

@article{deng2023mind2web,
  title={Mind2web: Towards a generalist agent for the web},
  author={Deng, Xiang and Gu, Yu and Zheng, Boyuan and Chen, Shijie and Stevens, Sam and Wang, Boshi and Sun, Huan and Su, Yu},
  journal={Advances in Neural Information Processing Systems},
  volume={36},
  pages={28091--28114},
  year={2023}
}

@article{wang2021codet5,
  title={Codet5: Identifier-aware unified pre-trained encoder-decoder models for code understanding and generation},
  author={Wang, Yue and Wang, Weishi and Joty, Shafiq and Hoi, Steven CH},
  journal={arXiv preprint arXiv:2109.00859},
  year={2021}
}

@article{yang2024swe,
  title={Swe-agent: Agent-computer interfaces enable automated software engineering},
  author={Yang, John and Jimenez, Carlos E and Wettig, Alexander and Lieret, Kilian and Yao, Shunyu and Narasimhan, Karthik and Press, Ofir},
  journal={Advances in Neural Information Processing Systems},
  volume={37},
  pages={50528--50652},
  year={2024}
}

@inproceedings{gao2023pal,
  title={Pal: Program-aided language models},
  author={Gao, Luyu and Madaan, Aman and Zhou, Shuyan and Alon, Uri and Liu, Pengfei and Yang, Yiming and Callan, Jamie and Neubig, Graham},
  booktitle={International Conference on Machine Learning},
  pages={10764--10799},
  year={2023},
  organization={PMLR}
}

@article{chen2022program,
  title={Program of thoughts prompting: Disentangling computation from reasoning for numerical reasoning tasks},
  author={Chen, Wenhu and Ma, Xueguang and Wang, Xinyi and Cohen, William W},
  journal={arXiv preprint arXiv:2211.12588},
  year={2022}
}

@article{Schnitzler2024MoreHopQAMT,
  title={MoreHopQA: More Than Multi-hop Reasoning},
  author={Julian Schnitzler and Xanh Ho and Jiahao Huang and Florian Boudin and Saku Sugawara and Akiko Aizawa},
  journal={ArXiv},
  year={2024},
  volume={abs/2406.13397},
  url={https://api.semanticscholar.org/CorpusID:270620469}
}

@article{li2023camel,
  title={Camel: Communicative agents for" mind" exploration of large language model society},
  author={Li, Guohao and Hammoud, Hasan and Itani, Hani and Khizbullin, Dmitrii and Ghanem, Bernard},
  journal={Advances in Neural Information Processing Systems},
  volume={36},
  pages={51991--52008},
  year={2023}
}

@inproceedings{wu2024autogen,
  title={Autogen: Enabling next-gen LLM applications via multi-agent conversations},
  author={Wu, Qingyun and Bansal, Gagan and Zhang, Jieyu and Wu, Yiran and Li, Beibin and Zhu, Erkang and Jiang, Li and Zhang, Xiaoyun and Zhang, Shaokun and Liu, Jiale and others},
  booktitle={First Conference on Language Modeling},
  year={2024}
}

@article{qian2023communicative,
  title={Communicative agents for software development},
  author={Qian, Chen and Cong, Xin and Yang, Cheng and Chen, Weize and Su, Yusheng and Xu, Juyuan and Liu, Zhiyuan and Sun, Maosong},
  journal={arXiv preprint arXiv:2307.07924},
  volume={6},
  number={3},
  pages={1},
  year={2023}
}

@inproceedings{hong2023metagpt,
  title={MetaGPT: Meta programming for a multi-agent collaborative framework},
  author={Hong, Sirui and Zhuge, Mingchen and Chen, Jonathan and Zheng, Xiawu and Cheng, Yuheng and Wang, Jinlin and Zhang, Ceyao and Wang, Zili and Yau, Steven Ka Shing and Lin, Zijuan and others},
  booktitle={The Twelfth International Conference on Learning Representations},
  year={2023}
}

@article{holt2023l2mac,
  title={L2MAC: Large language model automatic computer for extensive code generation},
  author={Holt, Samuel and Luyten, Max Ruiz and van der Schaar, Mihaela},
  journal={arXiv preprint arXiv:2310.02003},
  year={2023}
}

@inproceedings{zhuge2024gptswarm,
  title={Gptswarm: Language agents as optimizable graphs},
  author={Zhuge, Mingchen and Wang, Wenyi and Kirsch, Louis and Faccio, Francesco and Khizbullin, Dmitrii and Schmidhuber, J{\"u}rgen},
  booktitle={Forty-first International Conference on Machine Learning},
  year={2024}
}

@article{liu2023dynamic,
  title={Dynamic llm-agent network: An llm-agent collaboration framework with agent team optimization},
  author={Liu, Zijun and Zhang, Yanzhe and Li, Peng and Liu, Yang and Yang, Diyi},
  journal={arXiv preprint arXiv:2310.02170},
  year={2023}
}

@article{zhang2024g,
  title={G-designer: Architecting multi-agent communication topologies via graph neural networks},
  author={Zhang, Guibin and Yue, Yanwei and Sun, Xiangguo and Wan, Guancheng and Yu, Miao and Fang, Junfeng and Wang, Kun and Chen, Tianlong and Cheng, Dawei},
  journal={arXiv preprint arXiv:2410.11782},
  year={2024}
}

@article{wang2025anymac,
  title={AnyMAC: Cascading Flexible Multi-Agent Collaboration via Next-Agent Prediction},
  author={Wang, Song and Tan, Zhen and Chen, Zihan and Zhou, Shuang and Chen, Tianlong and Li, Jundong},
  journal={arXiv preprint arXiv:2506.17784},
  year={2025}
}

@article{zhou2023large,
  title={Large language model as a policy teacher for training reinforcement learning agents},
  author={Zhou, Zihao and Hu, Bin and Zhao, Chenyang and Zhang, Pu and Liu, Bin},
  journal={arXiv preprint arXiv:2311.13373},
  year={2023}
}

@inproceedings{ye2024uncertainty,
  title={Uncertainty regularized evidential regression},
  author={Ye, Kai and Chen, Tiejin and Wei, Hua and Zhan, Liang},
  booktitle={Proceedings of the AAAI Conference on Artificial Intelligence},
  volume={38},
  number={15},
  pages={16460--16468},
  year={2024}
}

@article{sensoy2018evidential,
  title={Evidential deep learning to quantify classification uncertainty},
  author={Sensoy, Murat and Kaplan, Lance and Kandemir, Melih},
  journal={Advances in neural information processing systems},
  volume={31},
  year={2018}
}

@inproceedings{gal2016dropout,
  title={Dropout as a bayesian approximation: Representing model uncertainty in deep learning},
  author={Gal, Yarin and Ghahramani, Zoubin},
  booktitle={international conference on machine learning},
  pages={1050--1059},
  year={2016},
  organization={PMLR}
}

@article{ovadia2019can,
  title={Can you trust your model's uncertainty? evaluating predictive uncertainty under dataset shift},
  author={Ovadia, Yaniv and Fertig, Emily and Ren, Jie and Nado, Zachary and Sculley, David and Nowozin, Sebastian and Dillon, Joshua and Lakshminarayanan, Balaji and Snoek, Jasper},
  journal={Advances in neural information processing systems},
  volume={32},
  year={2019}
}

@article{kuhn2023semantic,
  title={Semantic uncertainty: Linguistic invariances for uncertainty estimation in natural language generation},
  author={Kuhn, Lorenz and Gal, Yarin and Farquhar, Sebastian},
  journal={arXiv preprint arXiv:2302.09664},
  year={2023}
}

@article{kadavath2022language,
  title={Language models (mostly) know what they know},
  author={Kadavath, Saurav and Conerly, Tom and Askell, Amanda and Henighan, Tom and Drain, Dawn and Perez, Ethan and Schiefer, Nicholas and Hatfield-Dodds, Zac and DasSarma, Nova and Tran-Johnson, Eli and others},
  journal={arXiv preprint arXiv:2207.05221},
  year={2022}
}

@article{lin2023generating,
  title={Generating with confidence: Uncertainty quantification for black-box large language models},
  author={Lin, Zhen and Trivedi, Shubhendu and Sun, Jimeng},
  journal={arXiv preprint arXiv:2305.19187},
  year={2023}
}

@inproceedings{chen2024quantifying,
  title={Quantifying uncertainty in answers from any language model and enhancing their trustworthiness},
  author={Chen, Jiuhai and Mueller, Jonas},
  booktitle={Proceedings of the 62nd Annual Meeting of the Association for Computational Linguistics (Volume 1: Long Papers)},
  pages={5186--5200},
  year={2024}
}

@article{da2024llm,
  title={Llm uncertainty quantification through directional entailment graph and claim level response augmentation},
  author={Da, Longchao and Chen, Tiejin and Cheng, Lu and Wei, Hua},
  journal={arXiv preprint arXiv:2407.00994},
  year={2024}
}

@article{bai2023qwen,
  title={Qwen technical report},
  author={Bai, Jinze and Bai, Shuai and Chu, Yunfei and Cui, Zeyu and Dang, Kai and Deng, Xiaodong and Fan, Yang and Ge, Wenbin and Han, Yu and Huang, Fei and others},
  journal={arXiv preprint arXiv:2309.16609},
  year={2023}
}

@article{dubey2024llama,
  title={The llama 3 herd of models},
  author={Dubey, Abhimanyu and Jauhri, Abhinav and Pandey, Abhinav and Kadian, Abhishek and Al-Dahle, Ahmad and Letman, Aiesha and Mathur, Akhil and Schelten, Alan and Yang, Amy and Fan, Angela and others},
  journal={arXiv preprint arXiv:2407.21783},
  year={2024}
}

@inproceedings{gao-etal-2024-spuq,
    title = "{SPUQ}: Perturbation-Based Uncertainty Quantification for Large Language Models",
    author = "Gao, Xiang  and
      Zhang, Jiaxin  and
      Mouatadid, Lalla  and
      Das, Kamalika",
    editor = "Graham, Yvette  and
      Purver, Matthew",
    booktitle = "Proceedings of the 18th Conference of the European Chapter of the Association for Computational Linguistics (Volume 1: Long Papers)",
    month = mar,
    year = "2024",
    address = "St. Julian{'}s, Malta",
    publisher = "Association for Computational Linguistics",
    url = "https://aclanthology.org/2024.eacl-long.143/",
    doi = "10.18653/v1/2024.eacl-long.143",
    pages = "2336--2346"
}

@inproceedings{perros2017spartan,
  title={SPARTan: Scalable PARAFAC2 for large \& sparse data},
  author={Perros, Ioakeim and Papalexakis, Evangelos E and Wang, Fei and Vuduc, Richard and Searles, Elizabeth and Thompson, Michael and Sun, Jimeng},
  booktitle={Proceedings of the 23rd ACM SIGKDD International Conference on Knowledge Discovery and Data Mining},
  pages={375--384},
  year={2017}
}

@inproceedings{hou2024decomposing,
  title={Decomposing Uncertainty for Large Language Models through Input Clarification Ensembling},
  author={Hou, Bairu and Liu, Yujian and Qian, Kaizhi and Andreas, Jacob and Chang, Shiyu and Zhang, Yang},
  booktitle={International Conference on Machine Learning},
  pages={19023--19042},
  year={2024},
  organization={PMLR}
}

@article{catak2024uncertainty,
  title={Uncertainty quantification in large language models through convex hull analysis},
  author={Catak, Ferhat Ozgur and Kuzlu, Murat},
  journal={Discover Artificial Intelligence},
  volume={4},
  number={1},
  pages={90},
  year={2024},
  publisher={Springer}
}

@article{chen2025uncertainty,
  title={Uncertainty Quantification of Large Language Models through Multi-Dimensional Responses},
  author={Chen, Tiejin and Liu, Xiaoou and Da, Longchao and Chen, Jia and Papalexakis, Vagelis and Wei, Hua},
  journal={arXiv preprint arXiv:2502.16820},
  year={2025}
}

@article{kirchhof2025position,
  title={Position: Uncertainty quantification needs reassessment for large-language model agents},
  author={Kirchhof, Michael and Kasneci, Gjergji and Kasneci, Enkelejda},
  journal={arXiv preprint arXiv:2505.22655},
  year={2025}
}

@inproceedings{zhao2025uncertainty,
  title={Uncertainty Propagation on LLM Agent},
  author={Zhao, Qiwei and Li, Dong and Liu, Yanchi and Cheng, Wei and Sun, Yiyou and Oishi, Mika and Osaki, Takao and Matsuda, Katsushi and Yao, Huaxiu and Zhao, Chen and others},
  booktitle={Proceedings of the 63rd Annual Meeting of the Association for Computational Linguistics (Volume 1: Long Papers)},
  pages={6064--6073},
  year={2025}
}

@article{lu2024ai,
  title={The ai scientist: Towards fully automated open-ended scientific discovery},
  author={Lu, Chris and Lu, Cong and Lange, Robert Tjarko and Foerster, Jakob and Clune, Jeff and Ha, David},
  journal={arXiv preprint arXiv:2408.06292},
  year={2024}
}

@article{kim2024mdagents,
  title={Mdagents: An adaptive collaboration of llms for medical decision-making},
  author={Kim, Yubin and Park, Chanwoo and Jeong, Hyewon and Chan, Yik S and Xu, Xuhai and McDuff, Daniel and Lee, Hyeonhoon and Ghassemi, Marzyeh and Breazeal, Cynthia and Park, Hae W},
  journal={Advances in Neural Information Processing Systems},
  volume={37},
  pages={79410--79452},
  year={2024}
}

@article{tang2023medagents,
  title={Medagents: Large language models as collaborators for zero-shot medical reasoning},
  author={Tang, Xiangru and Zou, Anni and Zhang, Zhuosheng and Li, Ziming and Zhao, Yilun and Zhang, Xingyao and Cohan, Arman and Gerstein, Mark},
  journal={arXiv preprint arXiv:2311.10537},
  year={2023}
}

@article{fegade2022cora,
  title={The cora tensor compiler: Compilation for ragged tensors with minimal padding},
  author={Fegade, Pratik and Chen, Tianqi and Gibbons, Phillip and Mowry, Todd},
  journal={Proceedings of Machine Learning and Systems},
  volume={4},
  pages={721--747},
  year={2022}
}

@inproceedings{schenker2023parafac2,
  title={PARAFAC2-based coupled matrix and tensor factorizations},
  author={Schenker, Carla and Wang, Xiulin and Acar, Evrim},
  booktitle={ICASSP 2023-2023 IEEE International Conference on Acoustics, Speech and Signal Processing (ICASSP)},
  pages={1--5},
  year={2023},
  organization={IEEE}
}

@article{hendrycks2021measuring,
  title={Measuring mathematical problem solving with the math dataset},
  author={Hendrycks, Dan and Burns, Collin and Kadavath, Saurav and Arora, Akul and Basart, Steven and Tang, Eric and Song, Dawn and Steinhardt, Jacob},
  journal={arXiv preprint arXiv:2103.03874},
  year={2021}
}

@article{chen2021evaluating,
  title={Evaluating large language models trained on code},
  author={Chen, Mark and Tworek, Jerry and Jun, Heewoo and Yuan, Qiming and Pinto, Henrique Ponde De Oliveira and Kaplan, Jared and Edwards, Harri and Burda, Yuri and Joseph, Nicholas and Brockman, Greg and others},
  journal={arXiv preprint arXiv:2107.03374},
  year={2021}
}

@article{hendrycks2020measuring,
  title={Measuring massive multitask language understanding},
  author={Hendrycks, Dan and Burns, Collin and Basart, Steven and Zou, Andy and Mazeika, Mantas and Song, Dawn and Steinhardt, Jacob},
  journal={arXiv preprint arXiv:2009.03300},
  year={2020}
}

@article{Yang2024Qwen25TR,
  title={Qwen2.5 Technical Report},
  author={Qwen An Yang and Baosong Yang and Beichen Zhang and Binyuan Hui and Bo Zheng and Bowen Yu and Chengyuan Li and Dayiheng Liu and Fei Huang and Guanting Dong and Haoran Wei and Huan Lin and Jian Yang and Jianhong Tu and Jianwei Zhang and Jianxin Yang and Jiaxin Yang and Jingren Zhou and Junyang Lin and Kai Dang and Keming Lu and Keqin Bao and Kexin Yang and Le Yu and Mei Li and Mingfeng Xue and Pei Zhang and Qin Zhu and Rui Men and Runji Lin and Tianhao Li and Tingyu Xia and Xingzhang Ren and Xuancheng Ren and Yang Fan and Yang Su and Yi-Chao Zhang and Yunyang Wan and Yuqi Liu and Zeyu Cui and Zhenru Zhang and Zihan Qiu and Shanghaoran Quan and Zekun Wang},
  journal={ArXiv},
  year={2024},
  volume={abs/2412.15115},
  url={https://api.semanticscholar.org/CorpusID:274859421}
}

@article{bowman2015large,
  title={A large annotated corpus for learning natural language inference},
  author={Bowman, Samuel R and Angeli, Gabor and Potts, Christopher and Manning, Christopher D},
  journal={arXiv preprint arXiv:1508.05326},
  year={2015}
}

@article{zhao2024saup,
  title={SAUP: Situation Awareness Uncertainty Propagation on LLM Agent},
  author={Zhao, Qiwei and Zhao, Xujiang and Liu, Yanchi and Cheng, Wei and Sun, Yiyou and Oishi, Mika and Osaki, Takao and Matsuda, Katsushi and Yao, Huaxiu and Chen, Haifeng},
  journal={arXiv preprint arXiv:2412.01033},
  year={2024}
}

@book{maccartney2009natural,
  title={Natural language inference},
  author={MacCartney, Bill},
  year={2009},
  publisher={Stanford University}
}

@article{Yang2025Qwen3TR,
  title={Qwen3 Technical Report},
  author={An Yang and Anfeng Li and Baosong Yang and Beichen Zhang and Binyuan Hui and Bo Zheng and Bowen Yu and Chang Gao and Chengen Huang and Chenxu Lv and Chujie Zheng and Dayiheng Liu and Fan Zhou and Fei Huang and Feng Hu and Hao Ge and Haoran Wei and Huan Lin and Jialong Tang and Jian Yang and Jianhong Tu and Jianwei Zhang and Jianxin Yang and Jiaxin Yang and Jingren Zhou and Jingren Zhou and Junyan Lin and Kai Dang and Keqin Bao and Ke‐Pei Yang and Le Yu and Li-Chun Deng and Mei Li and Min Xue and Mingze Li and Pei Zhang and Peng Wang and Qin Zhu and Rui Men and Ruize Gao and Shi-Qiang Liu and Shuang Luo and Tianhao Li and Tianyi Tang and Wenbiao Yin and Xingzhang Ren and Xinyu Wang and Xinyu Zhang and Xuancheng Ren and Yang Fan and Yang Su and Yi-Chao Zhang and Yinger Zhang and Yu Wan and Yuqiong Liu and Zekun Wang and Zeyu Cui and Zhenru Zhang and Zhipeng Zhou and Zihan Qiu},
  journal={ArXiv},
  year={2025},
  volume={abs/2505.09388},
  url={https://api.semanticscholar.org/CorpusID:278602855}
}

@article{Kamath2025Gemma3T,
  title={Gemma 3 Technical Report},
  author={Gemma Team Aishwarya Kamath and Johan Ferret and Shreya Pathak and Nino Vieillard and Ramona Merhej and Sarah Perrin and Tatiana Matejovicova and Alexandre Ram'e and Morgane Rivi{\`e}re and Louis Rouillard and Thomas Mesnard and Geoffrey Cideron and Jean-Bastien Grill and Sabela Ramos and Edouard Yvinec and Michelle Casbon and Etienne Pot and Ivo Penchev and Gael Liu and Francesco Visin and Kathleen Kenealy and Lucas Beyer and Xiaohai Zhai and Anton Tsitsulin and R{\'o}bert Istvan Busa-Fekete and Alex Feng and Noveen Sachdeva and Benjamin Coleman and Yi Gao and Basil Mustafa and Iain Barr and Emilio Parisotto and David Tian and Matan Eyal and Colin Cherry and Jan-Thorsten Peter and Danila Sinopalnikov and Surya Bhupatiraju and Rishabh Agarwal and Mehran Kazemi and Dan Malkin and Ravin Kumar and David Vilar and Idan Brusilovsky and Jiaming Luo and Andreas Steiner and Abe Friesen and Abhanshu Sharma and Abheesht Sharma and Adi Mayrav Gilady and Adrian Goedeckemeyer and Alaa Saade and Alexander Kolesnikov and Alexei Bendebury and Alvin Abdagic and Amit Vadi and Andr'as Gyorgy and Andr{\'e} Susano Pinto and Anil Das and Ankur Bapna and Antoine Miech and Antoine Yang and Antonia Paterson and Ashish Shenoy and Ayan Chakrabarti and Bilal Piot and Boxi Wu and Bobak Shahriari and Bryce Petrini and Charlie Chen and Charline Le Lan and Christopher A. Choquette-Choo and Cj Carey and Cormac Brick and Daniel Deutsch and Danielle Eisenbud and Dee Cattle and Derek Cheng and Dimitris Paparas and Divyashree Shivakumar Sreepathihalli and Doug Reid and Dustin Tran and Dustin Zelle and Eric Noland and Erwin Huizenga and Eugene Kharitonov and Frederick Liu and Gagik Amirkhanyan and Glenn Cameron and Hadi Hashemi and Hanna Klimczak-Pluci'nska and Harman Singh and Harsh Mehta and Harshal Tushar Lehri and Hussein Hazimeh and Ian Ballantyne and Idan Szpektor and Ivan Nardini and Jean Pouget-Abadie and Jetha Chan and Joe Stanton and J. Michael Wieting and Jonathan Lai and Jordi Orbay and Joe Fernandez and Joshua Newlan and Junsong Ji and Jyotinder Singh and Kat Black and Kathy Yu and Kevin Hui and Kiran Vodrahalli and Klaus Greff and Linhai Qiu and Marcella Valentine and Marina Coelho and Marvin Ritter and Matt Hoffman and Matthew Watson and Mayank Chaturvedi and Michael Moynihan and Min Ma and Nabila Babar and Natasha Noy and Nathan Byrd and Nick Roy and Nikola Momchev and Nilay Chauhan and Oskar Bunyan and Pankil Botarda and Paul Caron and Paul Kishan Rubenstein and Phil Culliton and Philipp Schmid and Pier Giuseppe Sessa and Pingmei Xu and Piotr Stańczyk and Pouya Dehghani Tafti and Rakesh Shivanna and Renjie Wu and Renke Pan and Reza Ardeshir Rokni and Rob Willoughby and Rohith Vallu and Ryan Mullins and Sammy Jerome and Sara Smoot and Sertan Girgin and Shariq Iqbal and Shashir Reddy and Shruti Sheth and Siim P{\~o}der and Sijal Bhatnagar and Sindhu Raghuram Panyam and Sivan Eiger and Susan Zhang and Tianqi Liu and Trevor Yacovone and Tyler Liechty and Uday Kalra and Utku Evci and Vedant Misra and Vincent Roseberry and Vladimir Feinberg and Vlad Kolesnikov and Woohyun Han and Woosuk Kwon and Xi Chen and Yinlam Chow and Yuvein Zhu and Zichuan Wei and Zoltan Egyed and Victor Cotruta and Minh Giang and Phoebe Kirk and Anand Rao and Jessica Lo and Erica Moreira and Luiz Gustavo Martins and Omar Sanseviero and Lucas Gonzalez and Zach Gleicher and Tris Warkentin and Vahab S. Mirrokni and Evan Senter and Eli Collins and Joelle Barral and Zoubin Ghahramani and Raia Hadsell and Yossi Matias and D. Sculley and Slav Petrov and Noah Fiedel and Noam Shazeer and Oriol Vinyals and Jeffrey Dean and Demis Hassabis and Koray Kavukcuoglu and Cl{\'e}ment Farabet and Elena Buchatskaya and Jean-Baptiste Alayrac and Rohan Anil and Dmitry Lepikhin and Sebastian Borgeaud and Olivier Bachem and Armand Joulin and Alek Andreev and Cassidy Hardin and Robert Dadashi and L'eonard Hussenot},
  journal={ArXiv},
  year={2025},
  volume={abs/2503.19786},
  url={https://api.semanticscholar.org/CorpusID:277313563}
}

\clearpage
\appendix

\section{Code}

Code can be found at \url{https://github.com/tiejin98/MATU}.

\section{Theoretical Motivation: Reconstruction Loss as Semantic Variance}
\label{sec:appendix_theory}

To provide a clear motivation for our uncertainty quantification framework, we formalize the connection between our tensor-based approach and the classical statistical notion of variance.

In standard statistical theory, the variance of an ensemble measures the expected squared deviation of individual samples from the population mean (or consensus):
\begin{equation}
    \text{Var}(X) = \mathbb{E}\left[ \|X - \mu\|^2 \right]
\end{equation}

Our framework addresses the intractability of calculating this variance across discrete text trajectories by projecting the reasoning paths into a tensor space $\mathcal{X}$. Applying the PARAFAC2 (CP-2) decomposition yields a low-rank approximation $\hat{\mathcal{X}}$, which represents the \textit{expected semantic consensus} (the mean representation $\mu$) of the multi-agent system.

Consequently, the CP-2 reconstruction loss, formulated as the squared Frobenius norm of the residual:
\begin{equation}
    \mathcal{L} = \|\mathcal{X} - \hat{\mathcal{X}}\|_F^2 = \sum_{i,j,k} (x_{ijk} - \hat{x}_{ijk})^2
\end{equation}
is mathematically equivalent to the aggregated squared deviations from this consensus space. This motivates the use of \ours not as an ad-hoc heuristic, but as a principled instantiation of \textit{ensemble variance} for multi-agent latent spaces.

\section{Detailed Experimental Settings}

\subsection{Detailed Introduction to Datasets}
\label{sec:appendix_experimental_setting}

\begin{itemize}
    \item MATH~\citep{hendrycks2021measuring}: A dataset for \textbf{mathematical reasoning} that consists of challenging competition-level problems across algebra, geometry and number theory.
    \item MoreHopQA~\citep{Schnitzler2024MoreHopQAMT}:  A widely used question-answering dataset requiring \textbf{multi-hop text reasoning} over Wikipedia passages.
    \item MMLU~\citep{hendrycks2020measuring}: The Massive Multitask Language Understanding benchmark, covering 57 subjects. It assesses broad knowledge and problem-solving abilities, making it a strong indicator of \textbf{general-domain reasoning}. To avoid the overlap between the MATH dataset, when using MMLU, we exclude subjects about math.
    \item HumanEval~\citep{chen2021evaluating}: A code generation benchmark consisting of programming problems with unit tests. Models are required to synthesize correct and executable code solutions, and we provide a code environment for all multi-agent systems as a tool integration.
\end{itemize}

\subsection{Detailed Introduction to Baselines}
\label{sec:appendix_baseline}

As far as we know, we are the first method that targets the uncertainty quantification for MAS. To compare our method, we mainly adopt the existing methods for LLM or single-agent to multi-agent settings. In detail, we consider using Eigv(Agr)~\citep{lin2023generating}, which is the sum of eigenvalues for graph normalized Laplacian matrix and the graph is formed by the entailment matrix~\citep{bowman2015large} and P(true)~\citep{kadavath2022language}, which obtains the uncertainty by directly asking the LLM itself. For the Eigv(Agr), we use the final answer or every conversation to compute the entailment matrix, resulting in two different variants: Eigv(Agr)-answer and Eigv(Agr)-whole. Besides, we also use SAUP~\citep{zhao2024saup}, which is a white-box UQ method for a single agent by calculating the weighted sum of entropy for each step. We will treat the step from a different agent as each step in SAUP to transfer SAUP to a multi-agent setting. SAUP is originally designed for one trajectory, while we collect multiple trajectories. Therefore, we use SAUP-Single which uses the SAUP from the first trajectory, and SAUP-Multiple that uses the mean SAUP from all trajectories.

\section{Potential Explanation of Latent Component}
\label{sec:appendix_interpret}

To enhance the explainability of \ours, we can interpret the semantic meaning of the PARAFAC2 components by examining their corresponding factor vectors across the tensor modes (i.e., steps, agents, and runs). 

\subsection{Interpretation Protocol}
We interpret a given CP-2 latent component $r$ through the following steps:
\begin{enumerate}
    \item \textbf{Identify High-Loading Entities:} We first examine the factor vectors corresponding to the agents and runs modes. We identify the runs and agent roles with the highest scalar values (i.e., factor loadings, such as $u_{ir}$ for agents and $v_{jr}$ for runs) for component $r$. These scalar loadings act as quantitative indicators of how strongly a specific agent in a given run exhibits the latent pattern captured by component $r$.
    \item \textbf{Extract Semantic Meaning:} We then extract the top-weighted reasoning steps from the corresponding temporal/step factor to assign specific semantics to the component (e.g., identifying what specific textual logic correlates with the high loading).
\end{enumerate}

\subsection{Case Study: MATH Dataset}
We apply this protocol to the task \texttt{number\_theory\_60} from the MATH dataset. The original problem asks: \textit{``Suppose that $ABC$, where $A$, $B$, and $C$ are valid digits in base 4 and 9. What is the sum when you add all possible values of $A$, all possible values of $B$, and all possible values of $C$?''}

Applying our decomposition, we identified a specific latent component that aligns with a recurring failure pattern: a misunderstanding of the final summation target. While the assistant agent correctly finds all valid triples, it erroneously sums all digits across every solution instead of summing the distinct possible values as requested.

We verified this semantic error across multiple independent runs by examining the factor loadings for this specific error component:
\begin{itemize}
    \item \textbf{Incorrect Run 6:} The assistant’s factor loading spikes to 2.33. The corresponding generated log explicitly shows the flawed aggregation: \textit{``The sum of all A+B+C over these triples is 22.''}
    \item \textbf{Incorrect Run 3:} The scalar loading remains high at 1.83. The verbatim log confirms the exact same misinterpretation: \textit{``Sum of A's = 3+3+3+3 = 12.''}
    \item \textbf{Correct Run 2:} In stark contrast, the corresponding loading drops significantly to 0.57. Here, the assistant correctly interprets the summation rule and calculates the distinct values: \textit{``Possible A-values = \{3\} so $S_A = 3$ \dots Therefore $S_A + S_B + S_C = 10$.''}
\end{itemize}

Beyond individual instances, this semantic alignment is strongly validated by the overall numerical results. Across all runs for this task, the mean assistant factor loading for this specific component is 1.15 in incorrect runs, compared to only 0.50 in correct runs, yielding a clear $2.29\times$ separation ratio. This numerical contrast provides preliminary evidence that certain latent components (and their associated factor loadings) can localize recurring failure patterns at a statistical level. We leave the large-scale, automated semantic interpretation of latent factors across broader datasets as an important direction for future work.

\begin{table*}[t]
\centering
\small
\renewcommand{\arraystretch}{1.3} 
\begin{tabular}{p{0.18\linewidth} p{0.52\linewidth} p{0.22\linewidth}}
\toprule
\textbf{Case Info} & \textbf{Agent Reasoning Trajectories (Key Steps Only)} & \textbf{Uncertainty Quantification} \\ 
\midrule
\multirow{3}{=}{
    \textbf{Problem:} \newline 
    Find the distance between intersections of $y=x^2$ and $x+y=1$. \newline \newline
    \textbf{True Answer:} \newline 
    $\sqrt{10}$ \newline \newline
    \textbf{System Accuracy:} \newline 
    100\% (10/10 runs correct)
} 
& \textbf{Trajectory Type A: Direct \& Concise} \newline
$\bullet$ Determine intersection coordinates $\to$ Calculate distance \newline
\textit{\textcolor{gray}{Assessment: Ideal path, minimal token generation.}} 
& \multirow{3}{=}{
    \textbf{SAUP (Baseline):} \newline 
    \textcolor{red}{0.88 (High)} \newline 
    \scriptsize{\textit{Issue: High cumulative entropy from verbose steps.}} \newline
    
    \vspace{0.1cm}
    \textbf{Eigv-Whole (Baseline):} \newline 
    \textcolor{orange}{0.35 (Medium)} \newline 
    \scriptsize{\textit{Issue: Long contexts dilute NLI entailment accuracy.}} \newline

    \vspace{0.1cm}
    \textbf{MATU (Ours):} \newline 
    \textcolor{blue}{0.05 (Low)} \newline 
    \scriptsize{\textit{Result: Correctly aligns semantic intent across diverse paths.}}
} \\
\cmidrule{2-2}
& \textbf{Trajectory Type B: Self-Correction} \newline
$\bullet$ Determine coordinates $\to$ \textcolor{orange}{Correct y-axis calculation error} $\to$ Get correct coordinates $\to$ Calculate distance \newline
\textit{\textcolor{gray}{Assessment: Agent successfully recovers from an error.}} & \\
\cmidrule{2-2}
& \textbf{Trajectory Type C: Verbose (High Step Count)} \newline
$\bullet$ Determine coordinates $\to$ Get coordinates $\to$ \dots \textit{(intermediate steps)} \dots $\to$ Calculate distance \newline
\textit{\textcolor{gray}{Assessment: Logically identical to Type A, but higher step count increases cumulative entropy.}} & \\
\bottomrule
\end{tabular}
\caption{\textbf{Case Study on Mathematical Reasoning.} Despite diverse communication patterns, all agents consistently reach the correct solution ($\sqrt{10}$). Baselines like SAUP fail due to sensitivity to trajectory length (step count), and Eigv-Whole struggles with long-context entailment. MATU effectively disentangles surface-level variations from semantic stability, correctly assigning low uncertainty.}
\label{tab:case_study}
\end{table*}
\section{Case Study}
\label{sec:case_study}
To qualitatively demonstrate the robustness of MATU against the structural variability of multi-agent interactions, we analyze a representative example from the MATH dataset with qwen2.5, as illustrated in Table \ref{tab:case_study}. In detail, we have:

\noindent\textit{Question: ``What is the distance between the two intersections of $y=x^2$ and $x+y=1$?''} \\
\textit{Ground Truth: ``$\sqrt{10}$''}

In this experiment, we collected 10 independent reasoning trajectories. The multi-agent system demonstrated perfect performance, achieving a 100\% accuracy rate by deriving the correct answer $\sqrt{10}$ in all runs. However, the trajectories exhibited significant diversity in their communication patterns. While some runs produced concise and direct derivations (Type A), others involved self-correction mechanisms where agents identified and fixed calculation errors (Type B), or contained heavy steps (Type C), resulting in varying trajectory lengths with similar core logic when solving the problem. We report the normalized uncertainty values for all methods so that we might compare the uncertainty directly.

\paragraph{Analysis of Baselines.} 
Despite the consistency in the final outcome, baseline methods failed to accurately reflect the system's reliability. SAUP assigned a misleadingly high uncertainty score of 0.88. This false positive occurs because SAUP calculates uncertainty by accumulating entropy step-by-step. The heavy-step trajectories (Type C), despite being logically sound, contained more intermediate steps, which artificially inflated the cumulative entropy. Consequently, SAUP misinterpreted the surface-level verbosity, which is a byproduct of the communication topology, as semantic instability. Similarly, Eigv-Whole yielded a moderate uncertainty score of 0.35. This suggests that the NLI models used for entailment checking struggled to handle the long contexts and the noise introduced by self-correction steps, failing to fully recognize the logical entailment between the diverse reasoning paths.

\paragraph{Analysis of MATU.} 
In contrast, \textbf{MATU} correctly quantified the system's high reliability with a low uncertainty score of 0.05. By leveraging tensor decomposition on the reasoning embeddings, MATU effectively disentangles surface-level variations from the underlying semantic content. The tensor structure allows our method to align latent factors across trajectories of different lengths, recognizing that the corrective steps in Type B and the verbose explanations in Type C semantically converge to the same reasoning path as the concise Type A. This case highlights MATU's unique ability to filter out the noise caused by communication diversity, providing a more robust and holistic uncertainty measure for multi-agent systems.

\end{document}